\documentclass[sigconf]{acmart}

\usepackage{booktabs}
\usepackage{multirow}
\usepackage{graphicx}
\usepackage{amsmath}
\usepackage{balance}
\usepackage{enumitem}
\usepackage{array}
\usepackage{makecell}

\AtBeginDocument{%
  }

\copyrightyear{2026}
\acmYear{2026}
\setcopyright{cc}
\setcctype{by}

\acmConference[MM '26]
{Proceedings of the 34th ACM International Conference on Multimedia}
{November 10--14, 2026}
{Rio de Janeiro, Brazil}

\acmBooktitle{Proceedings of the 34th ACM International Conference
on Multimedia (MM '26), November 10--14, 2026,
Rio de Janeiro, Brazil}

\acmDOI{10.1145/3767308.3836001}
\acmISBN{979-8-4007-2213-4/2026/11}

\begin{document}

\title{DenseScout: Algorithm-System Co-design for Budgeted Tiny Object Selection on Edge Platforms}

\author{Zhouzhi Xiong}

\orcid{0009-0004-7891-1645}
\affiliation{%
  \institution{Zhejiang University}
  \city{Hangzhou}
  \country{China}}
\email{3220105776@zju.edu.cn}

\author{Zimo Zeng}
\orcid{0009-0008-8408-282X}
\affiliation{%
  \institution{Zhejiang University}
  \city{Hangzhou}
  \country{China}}
\email{zengzimo@zju.edu.cn}

\author{Yi Chen}
\orcid{0000-0001-5919-6413}
\affiliation{%
  \institution{Zhejiang University}
  \city{Hangzhou}
  \country{China}}
\email{morningone@126.com}

\author{Shuqi Xu}
\orcid{0009-0008-6942-3606}
\affiliation{%
  \institution{Zhejiang University}
  \city{Hangzhou}
  \country{China}}
\email{sqxu@zju.edu.cn}

\author{Yunfeng Yan}
\orcid{0000-0002-0939-5526}
\affiliation{%
  \institution{Zhejiang University}
  \city{Hangzhou}
  \country{China}}
\email{21210004@zju.edu.cn}

\author{Donglian Qi}
\correspondingauthor
\orcid{0000-0002-6535-2221}
\affiliation{%
  \institution{Zhejiang University}
  \city{Hangzhou}
  \country{China}}
\email{qidl@zju.edu.cn}

\renewcommand{\shortauthors}{Zhouzhi Xiong et al.}

\begin{abstract}
Deploying high-resolution tiny-object perception on edge platforms
requires not only accurate localization, but also selecting a small set
of informative patches under compute, transport, and latency constraints.
We study \emph{budgeted tiny-object selection}, where a frontend ranks
patch centers from a lightweight proxy and a downstream detector processes
only the selected regions. DenseScout is a 1.01M-parameter
deployment-oriented dense-response selector that removes detector-style
box regression and directly optimizes ranked patch-center prioritization.
Its contribution lies in the task-specific selector formulation, the alignment
among output representation, supervision, and decoding, and its joint
design with transport-aware execution and QoS-oriented evaluation.
Under unified protocols on VisDrone and DOTA, DenseScout provides stronger
low-budget recall than the evaluated detector-derived selectors;
controlled fixed-$K$ inspection experiments further demonstrate advantages
over selection-style proxy baselines. Cross-platform profiling on Jetson
Orin NX and RK3588 shows that deployable utility depends jointly on selector
quality, memory movement, and heterogeneous runtime realization. These
results support treating edge tiny-object perception as a
selection-and-deployment co-design problem rather than evaluating model
accuracy and runtime in isolation.
\end{abstract}



\maketitle

\section{Introduction}

High-resolution visual understanding is important in aerial monitoring, remote sensing, surveillance, and infrastructure inspection~\cite{power_inspection_apenergy_2025,power_inspection_eaai_2024,doi:10.1080/01431161.2023.2283900}. In such scenarios, targets may occupy only about $18\times14$ pixels in a $3840\times2160$ image, i.e., less than $0.05\%$ of the frame. This extreme scale sparsity makes reliable perception difficult, especially under strict patch and latency budgets.

\begin{figure}[!t]
    \centering
    \includegraphics[width=0.95\columnwidth]{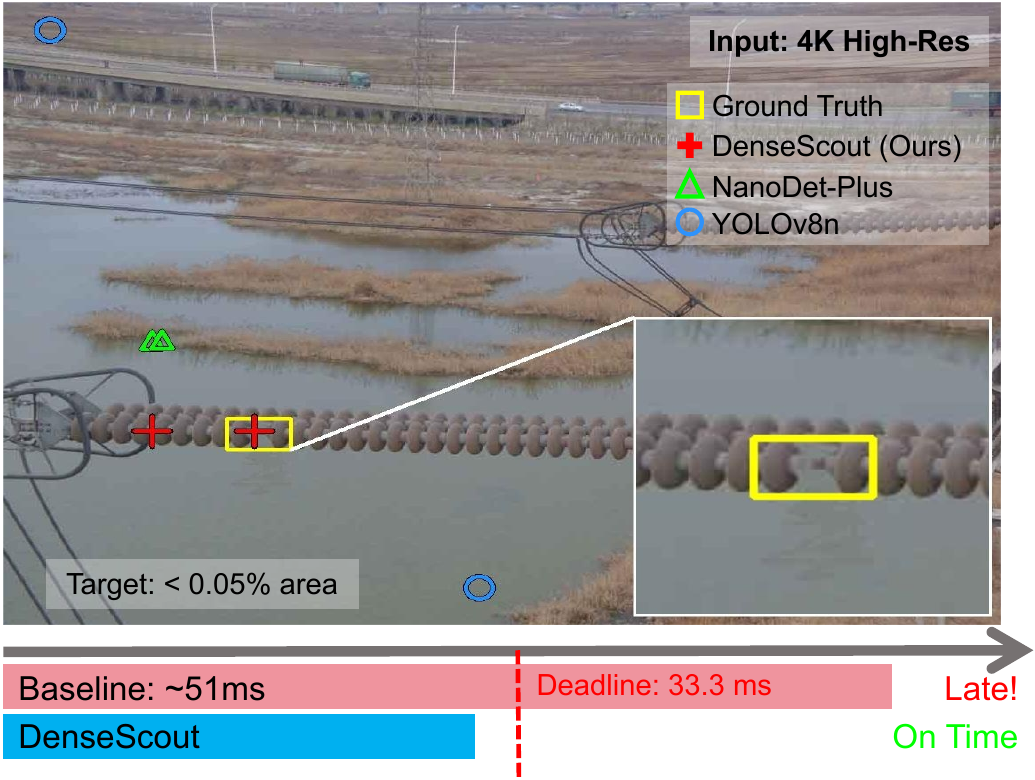}
    \caption{Motivating example of deployable tiny-object perception on the edge. In a 4K high-resolution frame, the target occupies less than 0.05\% of the image area and appears with weak visual saliency. Under a practical end-to-end deadline (33.3 ms), detector-based frontends may either miss the relevant region or exceed the latency budget, even if they are designed for generic object detection. In contrast, DenseScout preserves the target-containing patch under the same budget while remaining deadline-compliant. This example illustrates that offline detection or selection quality alone does not guarantee deployable utility.}
    \Description{A high-resolution inspection image with a tiny target, selected points from DenseScout and two detector baselines, and latency bars indicating whether each method meets a 33.3 ms deadline.}
    \label{fig:intro_motivation}
\end{figure}

A straightforward solution is to run detection on the full high-resolution image, but this is often impractical on edge platforms due to limited compute, memory bandwidth, and latency budgets~\cite{10.1109/TCSVT.2023.3297620,10.1145/3581783.3613785,10947590}. In many deployments, the system must first prioritize a small number of candidate patches and allocate downstream processing only to those regions. This leads to a \emph{budgeted tiny-object selection} problem: under a limited patch budget, the frontend should preserve as much useful tiny-target coverage as possible while remaining compatible with real-time execution.

Existing high-resolution pipelines often rely on detector-based frontends, coarse-to-fine strategies, patch-wise scanning, or tiling and offloading. Representative examples include~\cite{liu2023seeing,huang2025offnadir,vijayanarasimhan2010far,lee2014pertinent_patch,10.1109/INFOCOM42981.2021.9488843,10.1145/3680207.3765260,Girshick_2015_ICCV,He_2017_ICCV,Najibi_2019_ICCV,8578822}.
However, these designs are not well aligned with the setting studied here. Detector-based frontends are optimized for final localization rather than low-budget region prioritization, and offline metrics say little about whether selected regions remain useful under strict end-to-end deadlines on real hardware~\cite{10.1109/INFOCOM42981.2021.9488741,10.1145/3581783.3613785}. Figure~\ref{fig:intro_motivation} illustrates this gap. This mismatch motivates an \emph{algorithm--system co-design} view, in which selector quality and runtime realization are optimized jointly.

To address this gap, we formulate edge-side tiny-object perception as a joint \emph{selection-and-deployment} problem. DenseScout ranks patch centers under a fixed spatial budget; it does not perform final object recognition, which remains the responsibility of the downstream detector. The network itself is a lightweight realization of this interface rather than a new generic detector architecture. We further couple the selector with transport-aware execution and QoS-oriented evaluation on heterogeneous edge platforms.

The contributions are three-fold:
\begin{itemize}
    \item We formulate \textbf{budgeted ranked patch-center selection} as a deployment-facing frontend task, separating region prioritization from downstream recognition and connecting spatial coverage to deadline-valid utility.
    \item We develop \textbf{DenseScout}, a compact selector with \textbf{1.01M} parameters and \textbf{0.72} GFLOPs. Under unified protocols, it achieves strong practical-budget recall on VisDrone and DOTA and remains ahead of the evaluated detector-derived and selection-style proxy baselines.
    \item We develop a \textbf{transport-aware edge realization} and \textbf{QoS-oriented evaluation} that jointly account for coverage, budget, memory movement, and latency. Jetson Orin NX and RK3588 experiments show that selector quality must be considered together with heterogeneous runtime realization.
\end{itemize}

\paragraph{Code availability.}
The source-only alpha implementation of DenseScout is available at
\url{https://github.com/zhouer-bear/DenseScout}.

\section{Related Work}

\subsection{Tiny Object Detection and High-Resolution Visual Analytics}

Tiny object perception is challenging in aerial imagery, remote sensing, surveillance, and infrastructure inspection, where targets often occupy only a tiny fraction of the full image. Large-scale benchmarks such as VisDrone and DOTA have further highlighted the difficulty of high-resolution aerial perception, where objects are often tiny, densely distributed, weakly salient, and sometimes arbitrarily oriented~\cite{9573394,Xia_2018_CVPR,9560031}. Many studies improve detector performance through multi-scale fusion, attention mechanisms, feature enhancement, and lightweight detector design, such as PatchDetector, MDFFAM, DFE-DETR, Lino-YOLO, and SFPNet~\cite{ZHOU2024127715,xu_effective_2024,wu_dynamic_2025,11162630,11329099}. For aerial imagery, oriented-object detectors such as RoI Transformer explicitly model geometric rotation and improve localization on benchmarks like DOTA~\cite{Ding_2019_CVPR}. These methods are highly relevant to the tiny-object problem, but they primarily optimize the detector itself rather than an independent selector under an explicit patch budget.

Another line of work addresses high-resolution visual content through region proposals, sparse processing, patch-based reasoning, and adaptive context modeling~\cite{Tan_2019_ICCV,guan_region-based_2021,DBLP:journals/corr/abs-1911-08877,DBLP:journals/corr/abs-1805-09300,DBLP:journals/corr/abs-1904-08008,DBLP:journals/corr/abs-1811-08728,DBLP:journals/corr/abs-2106-10409}. These approaches support the idea that spatial regions should not be processed uniformly, but they typically operate on detector proposals, segmentation outputs, or adaptive patch context selection rather than unified top-$K$ tiny-object patch prioritization under a fixed budget.

DPR~\cite{zhang2023patchbasedselectionrefinementearly} is one of the few nearby methods that explicitly performs patch-wise selective processing before downstream detection. It first classifies image patches and then refines retained regions for later detection. Compared with our setting, however, DPR remains a multi-stage selection--refinement pipeline rather than a lightweight standalone selector optimized for fine-grained budgeted ranking. In contrast, our work studies a lightweight selector together with a transport-aware edge realization and a QoS-oriented evaluation framework to jointly assess coverage quality, runtime feasibility, and deadline-bounded utility.

\subsection{Resource-Aware Edge Visual Analytics}

Edge visual analytics has been widely studied under constraints of latency, bandwidth, memory movement, and energy. Existing efforts improve deployability through lightweight model design, model compression, runtime optimization, adaptive offloading, tile-level scheduling, and transmission-aware processing~\cite{10.1109/TCSVT.2023.3297620,10.1109/INFOCOM42981.2021.9488741,10.1109/INFOCOM42981.2021.9488843,10.1145/3581783.3613785,10.1145/3680207.3765260,10947590,DBLP:journals/corr/abs-2111-00902}. These techniques are effective for accelerating detector or backbone inference, but they usually treat the visual model as the primary optimization target. In high-resolution edge perception, however, deployable performance depends jointly on selector behavior, patch extraction, data transport, and downstream execution. This motivates a joint view of visual frontend design and runtime realization under strict patch and deadline budgets.

\subsection{QoS-Oriented Evaluation for Deployable Multimedia Perception}

Most visual perception studies report accuracy and runtime separately, such as AP, FPS, or latency. While informative, these metrics are insufficient for deployable edge perception because they do not directly capture whether useful target coverage is achieved within the required response window. Prior QoS-aware inference and edge analytics studies have highlighted the importance of latency-bounded and deadline-valid outputs in practical deployments~\cite{10.1109/INFOCOM42981.2021.9488741,10.1145/3581783.3613785,10.1145/3680207.3765260,10947590}. Recent systems have studied configurable detector ensembles for adaptive
accuracy--runtime--energy trade-offs%
~\cite{f1b8079e414c4f78be10f17c0699c818}, while ShaderNN demonstrates
how texture-native mobile GPU execution can reduce CPU--GPU data movement%
~\cite{XIE2025128628}. However, a unified QoS formulation tailored to budgeted tiny-object
patch selection remains underexplored. In our setting, utility depends
simultaneously on budget-efficient prioritization and deadline-bounded
execution. This motivates the QoS-oriented formulation adopted in this work.

\begin{figure*}[t]
    \centering
    \includegraphics[width=0.95\linewidth]{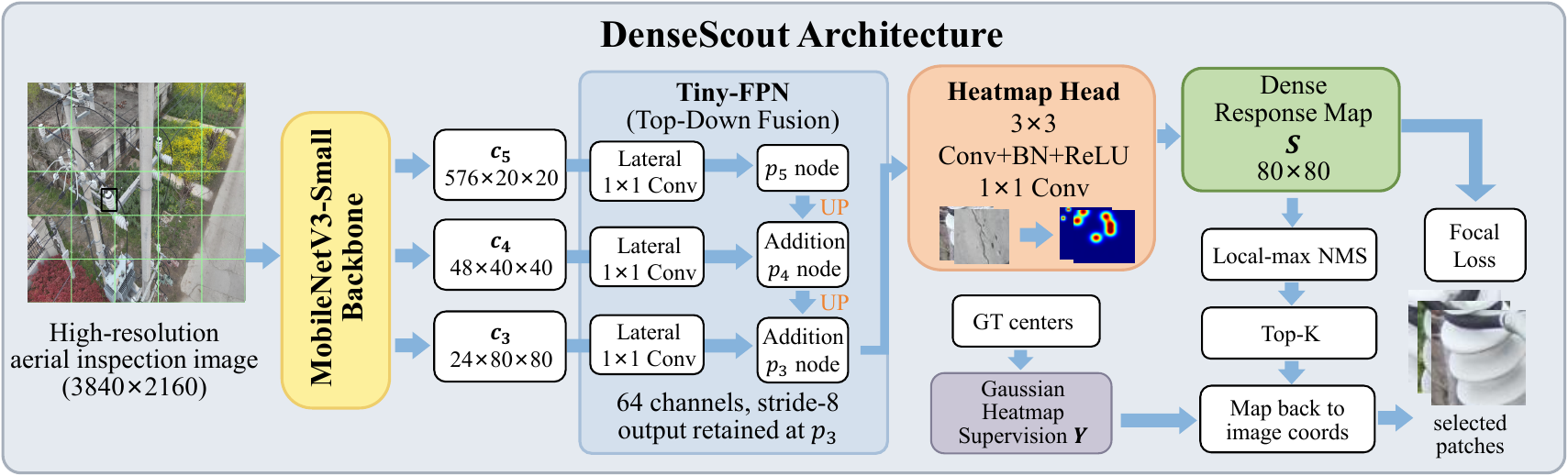}
    \caption{DenseScout architecture. A $640\times640$ proxy image is first processed by a MobileNetV3-Small backbone~\cite{Howard_2019_ICCV}, whose multi-scale features are fused by a Tiny-FPN to produce a stride-8 feature map $p_3$~\cite{Lin_2017_CVPR}. A lightweight heatmap head predicts an $80\times80$ dense response map. During training, ground-truth centers are converted into Gaussian heatmap supervision and optimized with a CenterNet-style focal loss. During inference, the response map is filtered by local-maximum suppression and truncated by Top-$K$ ranking, after which the selected centers are mapped back to image coordinates to generate fixed-size backend crops.}
    \Description{Block diagram of DenseScout, showing MobileNetV3-Small features, Tiny-FPN fusion, the 80 by 80 heatmap head, Gaussian supervision, local-maximum filtering, Top-K ranking, and mapping to backend crops.}
    \label{fig:densescout_arch}
\end{figure*}

In summary, prior studies improve detector accuracy, selective region processing, or runtime efficiency, but usually treat them in isolation. The focus of this work is distinct: budgeted tiny-object selection is framed as a deployment-oriented frontend problem, designed to jointly account for selection quality, runtime feasibility, and deadline-compliant utility.

\begin{figure*}[!t]
    \centering
    \includegraphics[width=0.95\linewidth]{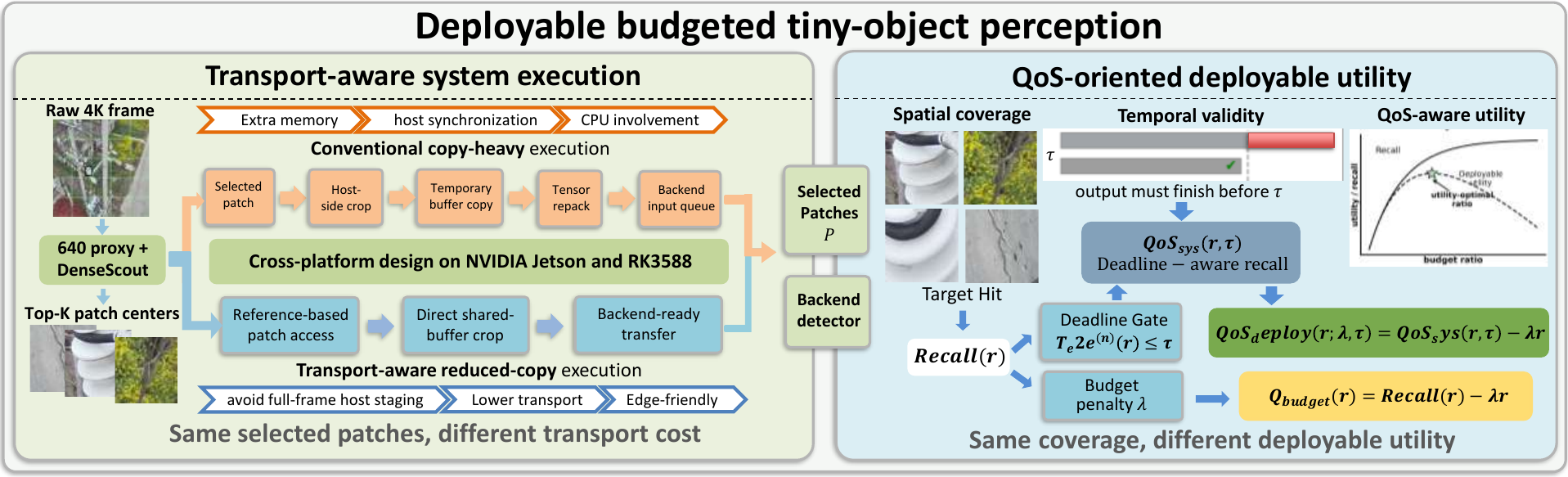}
    \caption{Deployable budgeted tiny-object perception with DenseScout. 
    Left: transport-aware system execution on edge platforms. Starting from a raw 4K frame, DenseScout first operates on a resized 640 proxy to produce Top-$K$ patch centers, after which the selected regions are executed through either a conventional copy-heavy path or a transport-aware Copy-Avoidance path before backend inference. 
    Right: QoS-oriented deployable utility. Spatial coverage, temporal validity, and budget-aware utility are jointly considered, showing that the same coverage can lead to different deployable utility under deadline and budget constraints.}
    \Description{Two-panel system diagram. The left panel compares copy-heavy and Copy-Avoidance patch execution on Jetson and RK3588. The right panel combines spatial coverage, temporal validity, and budget penalty into deployable QoS.}
    \label{fig:deployable_method}
\end{figure*}

\section{Method}

\subsection{DenseScout Model}

High-resolution tiny object perception is formulated as a \emph{budgeted patch selection} problem. 
Given an input image $I$, the frontend does not directly output final detection boxes. 
Instead, it selects a limited set of candidate patches
\begin{equation}
\mathcal{P}_K=\{p_1,p_2,\dots,p_K\}, \qquad \mathcal{P}_K \subset \mathcal{C}(I)
\label{eq:patch_budget}
\end{equation}

where $\mathcal{C}(I)$ is the candidate patch pool and $K$ is the maximum number of patch centers forwarded per frame to downstream processing. The frontend objective is to maximize tiny-target coverage under this budget; it does not solve the final detection task.

DenseScout is designed as a \emph{dense-response selector} rather than a detector-style proposal generator. 
For tiny objects, detector confidence and box regression are often unstable, whereas DenseScout predicts a dense response map on a low-resolution proxy image and ranks candidate regions according to local peak intensity. 
Its overall architecture is shown in Fig.~\ref{fig:densescout_arch}.

Given an input proxy image $I \in \mathbb{R}^{640\times640\times3}$, the backbone extracts a feature hierarchy
\begin{equation}
\{F_1,F_2,F_3,F_4,F_5\}=\mathcal{B}(I)
\label{eq:backbone_features}
\end{equation}

The last three stages are employed:
\begin{equation}
c_3 = F_3 \in \mathbb{R}^{24\times80\times80},\quad
c_4 = F_4 \in \mathbb{R}^{48\times40\times40},\quad
c_5 = F_5 \in \mathbb{R}^{576\times20\times20}
\label{eq:feature_shapes}
\end{equation}

After lateral $1\times1$ projection and top-down Tiny-FPN fusion, we obtain
\begin{equation}
p_5=\hat c_5,\qquad
p_4=\hat c_4+\operatorname{Up}(p_5),\qquad
p_3=\hat c_3+\operatorname{Up}(p_4)
\label{eq:tiny_fpn}
\end{equation}

where $\hat c_i$ denotes the 64-channel projected feature and $\operatorname{Up}(\cdot)$ denotes nearest-neighbor interpolation. 
A lightweight heatmap head then predicts
\begin{equation}
Z=\mathcal{H}(p_3), \qquad
S=\sigma(Z), \qquad
S\in[0,1]^{80\times80}
\label{eq:heatmap_output}
\end{equation}

Let $\mathcal{G}=\{g_m\}_{m=1}^{M}$ be the set of ground-truth
target centers, and let $(\bar{x}_m,\bar{y}_m)$ denote the center
of target $g_m$ in original-image pixels for an image of size
$W_0\times H_0$. We first normalize the coordinates as
$x_m=\bar{x}_m/W_0$ and $y_m=\bar{y}_m/H_0$, and then map them
to the stride-8 heatmap lattice:
\begin{equation}
\begin{aligned}
u_m &=
\left\lfloor
\min\!\left(79,\max(0,80x_m)\right)
\right\rfloor,\\
v_m &=
\left\lfloor
\min\!\left(79,\max(0,80y_m)\right)
\right\rfloor.
\end{aligned}
\label{eq:center_mapping}
\end{equation}
Thus, $u_m,v_m\in\{0,\ldots,79\}$ are the discrete lattice
coordinates used to place the Gaussian center.

A Gaussian heatmap supervision is then constructed as
\begin{equation}
Y(u,v)=
\max_{g_m\in\mathcal{G}}
\exp\!\left(
-\frac{(u-u_m)^2+(v-v_m)^2}{2\sigma^2}
\right)
\label{eq:gaussian_heatmap}
\end{equation}

DenseScout is trained with a CenterNet-style focal loss.
We first define the positive and negative per-location terms as
\begin{equation}
\ell_{uv}^{+}
=
(1-S_{uv})^2\log(S_{uv}),
\qquad
\ell_{uv}^{-}
=
(1-Y_{uv})^4S_{uv}^2\log(1-S_{uv}).
\label{eq:heat_loss}
\end{equation}

Let
\begin{equation}
\mathcal{P}=\{(u,v)\mid Y_{uv}=1\}
\label{eq:positive_set}
\end{equation}
denote the positive locations. The normalized training objective is
\begin{equation}
\mathcal{L}=
\begin{cases}
-\dfrac{1}{|\mathcal{P}|}
\left(
\displaystyle\sum_{(u,v)\in\mathcal{P}}\ell_{uv}^{+}
+
\displaystyle\sum_{(u,v)\notin\mathcal{P}}\ell_{uv}^{-}
\right),
& |\mathcal{P}|>0,\\[10pt]
-\displaystyle\sum_{u,v}\ell_{uv}^{-},
& |\mathcal{P}|=0.
\end{cases}
\label{eq:normalized_loss}
\end{equation}

At inference time, DenseScout first applies local-maximum filtering,
\begin{equation}
\hat S = S \odot \mathbf{1}\!\left[S=\operatorname{MaxPool}(S;k)\right]
\label{eq:local_max}
\end{equation}

and then ranks the retained peaks as
\begin{equation}
\{(u_i,v_i,s_i)\}_{i=1}^{N}=\operatorname{SortDesc}(\hat S)
\label{eq:sort_desc}
\end{equation}

Top-$K$ is applied only to valid peaks with positive score. If fewer than $K$ valid peaks exist, we use $K_{\mathrm{eff}}<K$ without zero-score padding:
\begin{equation}
\mathcal{P}_K=\{(u_i,v_i)\}_{i=1}^{K_{\mathrm{eff}}},
\qquad K_{\mathrm{eff}}=\min(K,N_{+}).
\label{eq:topk_patches}
\end{equation}

For an original image of size $W_0\times H_0$ under direct proxy resizing, a selected lattice location is mapped by
\begin{equation}
x_i^{p}=8u_i+4,\quad y_i^{p}=8v_i+4,\qquad
x_i^{0}=\frac{x_i^{p}W_0}{640},\quad
y_i^{0}=\frac{y_i^{p}H_0}{640}.
\label{eq:remap_coordinates}
\end{equation}
For letterbox preprocessing, the padding offset is removed before rescaling. Backend crops are clamped to the image boundary and zero-padded when necessary.

Each selected proxy center defines a fixed $64\times64$ evaluation cell for offline coverage accounting, and a proxy-space target center $(x_g^{p},y_g^{p})$ is counted as covered if
\begin{equation}
|x_i^{p}-x_g^{p}|\le 32,\qquad |y_i^{p}-y_g^{p}|\le 32.
\label{eq:coverage_condition}
\end{equation}
for at least one selected patch center. 
In the final training version, random horizontal flipping, HSV jitter, AdamW, and cosine annealing are additionally used to improve generalization without changing the inference graph. 
Overall, DenseScout removes detector-style box regression and complex decoding from the frontend, making it both more suitable for tiny-object selection and easier to deploy on edge platforms.

Two spatial units are distinguished. The $64\times64$ cell measures precise selector localization on the proxy lattice. The $640\times640$ crop is the downstream processing footprint expanded around the same center. Its coverage is therefore a loose spatial upper bound, not downstream detection success; end-task utility is measured separately by the closure benchmark. Unless stated otherwise, selector recall uses the $64\times64$ cell, whereas closure and deployment use $640\times640$ backend crops.

\subsection{System Realization on Edge Platforms}

DenseScout addresses \emph{what} to select, but deployable performance also depends on \emph{how} the selected patches are executed on real hardware. 
System realization is therefore considered part of the methodological approach. 
Its deployment logic and QoS-oriented evaluation are summarized in Fig.~\ref{fig:deployable_method}.

Given a raw $4$K frame $I^{4K}$, the deployment logic is illustrated in Fig.~\ref{fig:deployable_method}. 
A resized $640$ proxy is first processed by DenseScout to produce Top-$K$ patch centers. 
The selected regions are then extracted and delivered through either a conventional copy-heavy execution path or a transport-aware Copy-Avoidance execution path before backend inference and result aggregation.

The corresponding end-to-end latency is decomposed as
\begin{equation}
T_{\mathrm{e2e}}
=
T_{\mathrm{acq}}
+
T_{\mathrm{pre}}
+
T_{\mathrm{sel}}
+
T_{\mathrm{top}K}
+
T_{\mathrm{crop}}
+
T_{\mathrm{trans}}
+
T_{\mathrm{det}}
+
T_{\mathrm{post}}
\label{eq:e2e_latency}
\end{equation}

Here $T_{\mathrm{acq}}$ denotes frame/buffer access, $T_{\mathrm{pre}}$ proxy resize and normalization, $T_{\mathrm{sel}}$ selector inference, $T_{\mathrm{top}K}$ local-max decoding, $T_{\mathrm{crop}}$ patch extraction, $T_{\mathrm{trans}}$ inter-stage copy/synchronization/queueing, $T_{\mathrm{det}}$ backend inference, and $T_{\mathrm{post}}$ result aggregation.

Two timing boundaries separate model latency from pipeline overhead. \textit{Policy\_Only} measures accelerator submission-to-retrieval on a pre-allocated $640\times640\times3$ tensor, while \textit{Full\_FrontEnd} also includes frame intake, hardware-assisted resize/crop, and CPU post-processing. The conventional transport overhead is
\begin{equation}
T_{\mathrm{trans}} = T_{\mathrm{copy}} + T_{\mathrm{sync}} + T_{\mathrm{queue}}.
\label{eq:transport_latency}
\end{equation}
Our Copy-Avoidance path minimizes redundant user-space staging rather than claiming universally copy-free execution.

On Jetson Orin NX, DenseScout uses TensorRT FP16, reusable device buffers, and CUDA streams; host--device movement and stream synchronization dominate the non-inference overhead. On RK3588, it uses RKNN FP16 with RGA/DMA-assisted preprocessing; toolchain compatibility, tensor stride, cache coherence, and buffer layout are the main implementation constraints. The RK3588 selector time is decomposed as
\begin{equation}
T_{\mathrm{sel}} = T_{\mathrm{NPU}} + T_{\mathrm{CPU\text{-}post}},
\label{eq:selector_latency}
\end{equation}
where the second term contains local-max extraction and Top-$K$ decoding. To characterize runtime feasibility, we monitor throughput and tail behavior. For a stream of $N_f$ frames,

\begin{equation}
\mathrm{FPS}=\frac{1000N_f}{\sum_{n=1}^{N_f}T_{\mathrm{e2e,ms}}^{(n)}},
\qquad
\mathrm{Jitter}=p99-p50
\label{eq:fps_jitter}
\end{equation}
When latency is measured in milliseconds, the equivalent FPS form is used as above.

Overall, the system realization is not a simple wrapper around DenseScout: its single-heatmap output makes frontend decoding lightweight, while Copy-Avoidance, probes, and platform-aware runtime coordination ensure that the selector’s spatial advantage can be translated into deployable utility.

\subsection{QoS-Oriented Formulation and Metrics}

After defining DenseScout and its edge-side realization, a unified QoS formulation for \emph{deployable utility} is introduced. 
The core idea is that a selector is useful only if it achieves strong target coverage under limited budget consumption and if this advantage can still be realized within the system deadline. 
The relationship among the QoS terms is also illustrated in Fig.~\ref{fig:deployable_method}, whose right panel summarizes the deployable utility view.

Let $\mathcal{G}=\{g_1,\dots,g_M\}$ be the set of ground-truth tiny targets and let $\mathcal{P}$ denote the selected patch set. 
The budget ratio is defined as
\begin{equation}
r(\mathcal{P})=\frac{\left|\bigcup_{p\in\mathcal{P}}p\right|}{HW}
\label{eq:budget_ratio}
\end{equation}

where $H\times W$ is the full image size. 
A target $g\in\mathcal{G}$ is counted as covered if its center lies inside at least one selected patch:
\begin{equation}
\mathrm{hit}(g,\mathcal{P})=
\begin{cases}
1, & \exists\, p\in\mathcal{P}\ \text{s.t.}\ g\in p,\\
0, & \text{otherwise}
\end{cases}
\label{eq:hit_definition}
\end{equation}

The corresponding coverage recall is
\begin{equation}
\mathrm{Recall}(r)=
\frac{1}{|\mathcal{G}|}
\sum_{g\in\mathcal{G}}
\mathbf{1}\!\left[\mathrm{hit}(g,\mathcal{P})\right]
\label{eq:coverage_recall}
\end{equation}

For frame $n$, let $\mathcal{G}_n$ denote its ground-truth target
set, and let $\mathcal{P}_n(r)$ denote the patch set selected under
budget ratio $r$. Based on the coverage recall in
Eq.~\eqref{eq:coverage_recall}, we define three complementary QoS
terms.

First, budget-aware utility penalizes the retained spatial area:
\begin{equation}
\mathrm{QoS}_{\mathrm{budget}}(r;\lambda)
=
\mathrm{Recall}(r)-\lambda r ,
\label{eq:qos_budget}
\end{equation}
where $\lambda$ is a deployment-side budget price.

Second, system-level QoS measures deadline-valid target coverage:
\begin{equation}
\mathrm{QoS}_{\mathrm{sys}}(r,\tau)
=
\frac{
\sum_{n=1}^{N_f}
\sum_{g\in\mathcal{G}_n}
\mathbf{1}\!\left[
\mathrm{hit}\!\left(g,\mathcal{P}_n(r)\right)
\right]
\mathbf{1}\!\left[
T_{\mathrm{e2e}}^{(n)}(r)\le\tau
\right]
}{
\sum_{n=1}^{N_f}|\mathcal{G}_n|
},
\label{eq:qos_sys}
\end{equation}
where the deadline gate is applied at the frame level before target
hits are aggregated.

Finally, deployable utility jointly accounts for deadline-valid
coverage and spatial budget:
\begin{equation}
\mathrm{QoS}_{\mathrm{deploy}}(r;\lambda,\tau)
=
\mathrm{QoS}_{\mathrm{sys}}(r,\tau)-\lambda r .
\label{eq:qos_deploy}
\end{equation}

Thus, a spatially covered target contributes to
$\mathrm{QoS}_{\mathrm{sys}}$ only when its corresponding frame
satisfies $T_{\mathrm{e2e}}^{(n)}(r)\le\tau$.
Recall and $r$ are represented in $[0,1]$ when computing these
terms. The coefficient $\lambda$ is not learned; it assigns increasing
cost to additional selected area, transport, and backend crop
computation. The threshold $\tau$ is the application deadline
(e.g., 33.3\,ms or 15\,ms). The coefficient $\lambda$ controls the coverage--budget trade-off,
with larger values assigning a higher penalty to retained area.

This QoS family answers not only which method achieves higher recall, but also which one reaches its best utility at a smaller budget ratio and whether that operating point remains valid under real deployment constraints.

\begin{table*}[!t]
\centering
\caption{Unified comparison of frontend selectors and full-image detectors. Full-image detector accuracy is only applicable to conventional detectors evaluated under the standard image-level detection protocol, while DenseScout and DPR are budget-aware frontend selection methods rather than standalone detectors. Efficiency and Recall@Ratio are reported under the same patch-selection setting.}
\label{tab:unified_main_results}
\setlength{\tabcolsep}{3.8pt}
\renewcommand{\arraystretch}{0.60}
\small
\begin{tabular}{llcccc|ccc|ccc}
\toprule
\multirow{2}{*}{Category} & \multirow{2}{*}{Method}
& \multicolumn{2}{c}{VisDrone-val Detector Accuracy}
& \multicolumn{2}{c|}{Efficiency}
& \multicolumn{3}{c|}{VisDrone Recall@Ratio (\%)}
& \multicolumn{3}{c}{DOTA Recall@Ratio (\%)} \\
\cmidrule(lr){3-4} \cmidrule(lr){5-6} \cmidrule(lr){7-9} \cmidrule(lr){10-12}
& & AP$_{50}$ & AP$_{50:95}$ & Params (M) & GFLOPs & 1\% & 2\% & 4\% & 1\% & 2\% & 4\% \\
\midrule
\multirow{2}{*}{Frontend selectors}
& DenseScout    & --     & --     & \textbf{1.01} & \textbf{0.72} & \textbf{35.16} & \textbf{51.15} & \textbf{68.93} & \textbf{28.52} & \textbf{38.20} & \textbf{48.40} \\
& DPR           & --     & --     & 65.17         & 28.34         & 26.47          & 34.29          & 42.74          & 6.64           & 11.14          & 15.67          \\
\midrule
\multirow{5}{*}{Full-image detectors}
& YOLO11n       & 0.342  & 0.198  & 2.59          & 3.23          & 9.14           & 18.76          & 36.52          & 12.49          & 18.92          & 27.66          \\
& YOLOv8n       & 0.505  & 0.314  & 3.01          & 4.10          & 9.43           & 18.57          & 38.16          & 10.78          & 17.57          & 25.96          \\
& NanoDet-Plus  & 0.240  & 0.140  & 4.18          & 1.85          & 9.35           & 18.44          & 36.21          & 18.39          & 25.79          & 37.92          \\
& RT-DETR-light & 0.583  & 0.378  & 32.83         & 54.02         & 8.45           & 17.23          & 35.85          & 13.69          & 20.11          & 28.93          \\
& RetinaNet-R50 & 0.0653 & 0.0419 & 32.36         & $\sim$145     & 11.66          & 21.08          & 40.55          & 16.07          & 22.89          & 32.38          \\
\bottomrule
\end{tabular}
\end{table*}

\section{Experimental Results}
\label{sec:results}

\subsection{Datasets}
\label{subsec:datasets}

Different datasets are employed for complementary evaluation purposes. VisDrone~\cite{9573394} and DOTA~\cite{Xia_2018_CVPR,9560031} are used for the main offline evaluation of budgeted tiny-object selection, where we measure Recall@Ratio under unified patch-budget constraints. For deployment-oriented evaluation on real edge boards, we use InsPLAD~\cite{doi:10.1080/01431161.2023.2283900}, a UAV-based infrastructure inspection dataset whose high-resolution imagery, sparse tiny targets, and realistic inspection context match the targeted edge deployment scenarios.

VisDrone and DOTA benchmark generic budgeted selection, while InsPLAD provides a realistic edge-inspection workload. To close the metric gap across datasets, we also evaluate selector quality on InsPLAD: DenseScout obtains 92.28/92.38/92.46 Recall@1/2/4\% and 77.89 Recall@$K{=}9$, compared with DPR's 35.09/48.07/58.62 and 23.36.

\subsection{Unified Selector Protocol}
\label{subsec:selector_protocol}

Detector-derived baselines use post-NMS boxes and final confidence. Each $80\times80$ cell whose center falls inside a scaled box inherits that confidence, and overlapping boxes are merged by the maximum score before ranking. Only positive-score candidates are retained; if fewer than $K$ exist, $K_{\mathrm{eff}}<K$ is used without padding. A $K{=}9$ audit yields mean $K_{\mathrm{eff}}=9.0$ and $\Pr(K_{\mathrm{eff}}<K)=0$ for all detector-derived baselines.

\paragraph{Signal-level conversion control.}
To isolate the effects of score representation and decoding, we conduct
a separate control study on the VisDrone validation split
(548 images and 38{,}791 targets). All methods share an
$80\times80$ candidate grid, where 0.5/1/2\% correspond to
Top-32/64/128 cells. For DenseScout, this control uses the raw response
map without local-max suppression ($k=1$), and is therefore distinct
from the main Recall@Ratio protocol in
Table~\ref{tab:unified_main_results}. DenseScout obtains
62.10/77.90/89.11 Recall@0.5/1/2\%, while CenterNet heatmap Top-$K$
obtains 61.64/77.32/88.97. CenterNet covered-cell and FCOS
class$\times$centerness signals obtain 27.00/38.67/53.47 and
22.05/35.50/53.86, respectively. These results show that raw heatmap
ranking is a strong selector signal; our contribution is not Top-$K$
itself, but the task-specific selector design and its
deployment-oriented formulation.

\subsection{Main Recall--Budget Performance}
\label{subsec:main_recall_budget}

\begin{figure}[t]
    \centering
    \includegraphics[width=0.48\textwidth]{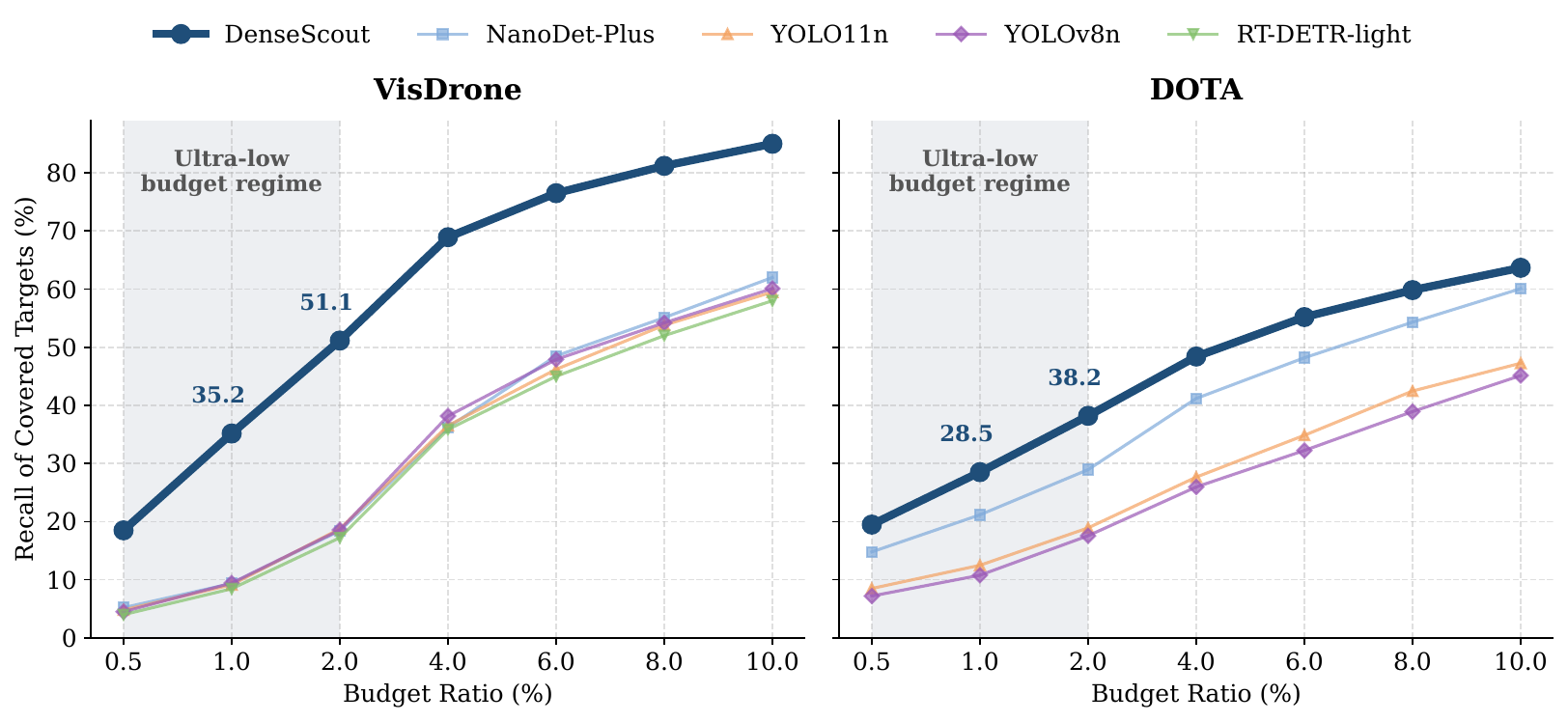}
    \caption{Recall@Ratio curves on VisDrone and DOTA. DenseScout consistently provides the best target coverage under tight patch-budget constraints, with the clearest margin in the ultra-low and practical budget regime (1\%--4\%).}
    \Description{Two Recall-at-Ratio curves for VisDrone and DOTA. DenseScout is above detector-derived baselines throughout the practical low-budget region.}
    \label{fig:recall_budget_dual_main}
\end{figure}

DenseScout is initially compared against detector-based and proposal-based baselines under the same budgeted patch-selection protocol, with all frontends converted to ranked patch centers using the same budget and selection rules.
Figure~\ref{fig:recall_budget_dual_main} shows Recall@Ratio on VisDrone and DOTA. DenseScout consistently provides the strongest target coverage under tight budgets, with the clearest advantage in the practical low-budget regime of 1\%--4\%.

To complement the trend plots, Table~\ref{tab:unified_main_results} summarizes conventional detector accuracy, model efficiency, and practical-budget Recall@Ratio. This view is intentionally comprehensive: some baselines have competitive full-image detector mAP, but still underperform as low-budget patch selectors. DenseScout remains the strongest method in practical-budget recall on both datasets while also being the lightest model in terms of parameters and FLOPs.

As a recent edge-oriented control,
YOLO26n~\cite{jocher2026ultralyticsyolo26} obtains 9.07/ 18.19/ 37.83 Recall@1/ 2/ 4\% on VisDrone under the same detector-to-selector protocol, compared with DenseScout's 35.16/ 51.15/ 68.93. This indicates that detector efficiency does not necessarily translate into effective fixed-budget patch selection.

To verify that better patch selection also improves end-task utility, we further add a patch-based detection closure benchmark on VisDrone-2019 val. Under a fixed budget of $K{=}9$, each frontend selects nine $640\times640$ patches, which are then processed by a frozen YOLOv8n backend and merged by class-aware global NMS. As shown in Table~\ref{tab:closure_benchmark}, DenseScout outperforms
DPR under both the custom closure evaluator and the official COCO
bbox evaluator.

\begin{table}[!t]
\centering
\caption{Patch-based detection closure benchmark on VisDrone-2019
val with a frozen YOLOv8n backend under a strict $K{=}9$ patch
budget. Closure R@9 and Closure AP$_{50}^{\mathrm{11pt}}$ are computed
by our class-aware merged-detection evaluator at IoU$\,{=}\,0.5$;
COCO AP$_{50}$ and AP$_{50:95}$ are computed using
\texttt{pycocotools.COCOeval} over the merged detections.}
\label{tab:closure_benchmark}
\setlength{\tabcolsep}{4.0pt}
\renewcommand{\arraystretch}{0.72}
\footnotesize
\begin{tabular}{lcccc}
\toprule
Frontend
& Closure R@9
& Closure AP$_{50}^{\mathrm{11pt}}$
& COCO AP$_{50}$
& COCO AP$_{50:95}$ \\
\midrule
DenseScout
& \textbf{79.4}
& \textbf{59.2}
& \textbf{41.75}
& \textbf{24.43} \\
DPR
& 73.4
& 55.4
& 38.62
& 22.21 \\
\bottomrule
\end{tabular}
\end{table}

Several observations follow. First, strong full-image detector accuracy does not imply strong low-budget patch selection: for example, RT-DETR-light achieves the highest detector mAP among the generic detector baselines, yet remains clearly behind DenseScout in practical-budget recall. Second, DenseScout does not trade efficiency for coverage; it delivers both the best recall and the lowest model cost, making it a better-matched frontend for budget-constrained target coverage.

\subsection{Selection-Style Baselines}
The following selection-style comparison and granular ablation use the
original fixed-K inspection protocol on the resolved labeled
Lab\_Dataset validation subset, containing 371 images and 975 point
targets. These fixed-$K$ results are not directly comparable to the
VisDrone/DOTA Recall@Ratio results.
\label{subsec:selection_baselines}

To complement detector-derived baselines, we evaluate four selection-style proxy/lite implementations under the original fixed-$K$ tiny-fault protocol. These are not official reproductions; they test whether focus-, mask-, zoom-, or dynamic-head-style proposal mechanisms explain the gain.

\begin{table}[t]
\centering
\caption{Selection-style comparison at $K=9$ on the labeled
Lab\_Dataset validation subset under the fixed-K inspection protocol. The proxy/lite baselines are not official reproductions.}
\label{tab:selection_style}
\setlength{\tabcolsep}{13.2pt}
\renewcommand{\arraystretch}{0.60}
\footnotesize
\begin{tabular}{lc@{\hspace{1.4em}}lc}
\toprule
Method & R@9 & Method & R@9 \\
\midrule
\textbf{DenseScout} & \textbf{80.00} & GigaDet-DH-lite & 51.79 \\
MaskProposer & 56.10 & AdaZoom-lite & 50.05 \\
FocusGrid & 42.87 & DPR & 11.28 \\
\bottomrule
\end{tabular}
\end{table}

The strongest added baseline, MaskProposer, reaches 56.10, remaining 23.90 points below DenseScout.

\subsection{Qualitative Comparison}

\begin{figure}[!t]
    \centering
    \includegraphics[width=0.45\textwidth]{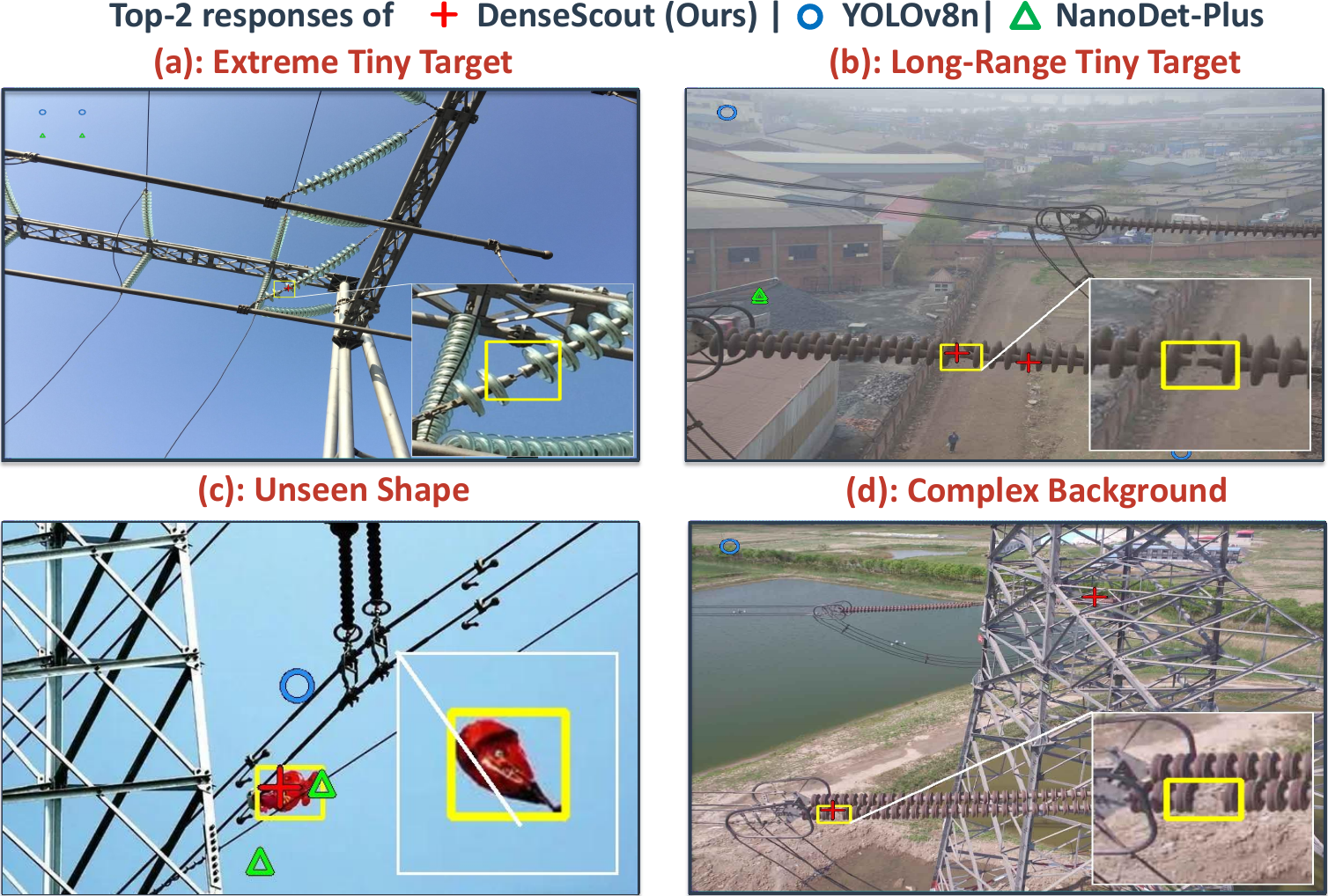}
    \caption{Top-ranked responses on four challenging cases.
    Yellow boxes mark ground truth; red crosses, blue circles, and
    green triangles denote DenseScout, YOLOv8n, and NanoDet-Plus,
    respectively.}
    \Description{Four qualitative inspection examples comparing the top-ranked responses of DenseScout, YOLOv8n, and NanoDet-Plus around tiny or visually ambiguous targets.}
    \label{fig:qualitative_cases}
\end{figure}

Figure~\ref{fig:qualitative_cases} compares the top-ranked responses of different frontends on four challenging cases, including extremely small targets, long-range tiny target, unseen shape, and complex background clutter. DenseScout places its highest-ranked responses more consistently near the ground-truth tiny targets, whereas detector-based frontends are more easily distracted by background structures or irrelevant salient regions.

\subsection{Ablation Study}
\label{subsec:ablation}

We isolate architecture, training, decoding, and supervision under the same fixed-$K$ protocol. Full DenseScout with local-max kernel $k=7$ reaches 80.00 Recall@$K{=}9$.

\begin{table}[!t]
\centering
\caption{Granular ablation on the labeled Lab\_Dataset validation
subset under the fixed-K inspection protocol.}
\label{tab:ablation_main}
\setlength{\tabcolsep}{12.8pt}
\renewcommand{\arraystretch}{0.80}
\footnotesize
\begin{tabular}{lclc}
\toprule
Variant & R@9 & Variant & R@9 \\
\midrule
Shallow $80\times80$ & 57.03 & Raw Top-$K$ & 68.21 \\
Tiny-FPN default & 74.26 & Binary BCE disk & 50.36 \\
Augmentation only & 79.90 & Gaussian MSE & 43.79 \\
\textbf{DenseScout full} & \textbf{80.00} & Default $k=7$ & \textbf{80.00} \\
\bottomrule
\end{tabular}
\end{table}

The results show that high-resolution features alone are insufficient; Tiny-FPN, augmentation, local-max decoding, and Gaussian-focal supervision jointly account for the final performance.

The practical operating region is the low-budget regime, where transport, memory movement, and downstream throughput restrict the retained area. DenseScout provides the steepest recall gain per unit budget in this regime on both datasets.

\begin{table*}[!t]
\centering
\caption{Cross-platform runtime and QoS-aware recall at the practical budget $K=9$ on InsPLAD-based deployment workloads. We report raw recall, deadline-aware recall, latency statistics, and tail jitter under different transport settings. Tail jitter is defined as $p99-p50$.}

\label{tab:cross_platform_qos_runtime}
\setlength{\tabcolsep}{3.2pt}
\renewcommand{\arraystretch}{0.60}
\small
\begin{tabular}{ll|cccccc|cccccc}
\toprule
& & \multicolumn{6}{c|}{Jetson Orin NX} & \multicolumn{6}{c}{RK3588} \\
\cmidrule(lr){3-8} \cmidrule(lr){9-14}
Method & Transport
& Raw R@9 & QoS@33.3ms & QoS@15ms & p50 & p99 & Jitter
& Raw R@9 & QoS@33.3ms & QoS@15ms & p50 & p99 & Jitter \\
\midrule
DenseScout & Copy
& 78.12 & 78.12 & 77.91 & 3.05 & 3.13 & 0.08
& 76.90 & 76.90 & 0.00 & 24.98 & 26.29 & 1.31 \\

DenseScout & Copy-Avoidance
& \textbf{78.12} & \textbf{78.12} & \textbf{78.12} & \textbf{2.98} & \textbf{3.06} & \textbf{0.08}
& \textbf{76.90} & \textbf{76.90} & 0.00 & \textbf{24.97} & \textbf{26.18} & \textbf{1.21} \\

NanoDet-Plus & Copy
& 40.63 & 40.63 & 40.43 & 6.79 & 6.87 & 0.08
& 44.28 & 0.00 & 0.00 & 49.17 & 71.62 & 22.45 \\

NanoDet-Plus & Copy-Avoidance
& 40.63 & 40.63 & 40.63 & 6.70 & 6.79 & 0.09
& 44.28 & 0.00 & 0.00 & 48.57 & 72.26 & 23.69 \\

YOLOv8n & Copy
& 59.68 & 59.68 & 59.47 & 6.20 & 6.32 & 0.12
& -- & 0.00 & 0.00 & 60.75 & 71.07 & 10.32 \\

YOLOv8n & Copy-Avoidance
& 59.68 & 59.68 & 59.68 & 6.04 & 6.14 & 0.10
& -- & 0.00 & 0.00 & 62.56 & 74.35 & 11.79 \\
\bottomrule
\end{tabular}
\end{table*}

\section{Deployment and System Analysis}
\label{sec:system_analysis}

Beyond offline recall, we evaluate whether DenseScout preserves its advantage under realistic edge deployment constraints. Unless otherwise specified, the deployment experiments in this section are conducted on InsPLAD~\cite{doi:10.1080/01431161.2023.2283900}, which provides a practical UAV-based inspection workload for board-level runtime and QoS analysis. We focus on cross-platform latency--QoS trade-offs, deadline-aware recall, and runtime composition under different transport settings. DPR is included in offline selection and closure evaluation. We additionally profile its selector stage on Jetson Orin NX using TensorRT FP16 with Top-$K$ post-processing: 14.38\,ms mean and 15.33\,ms $p_{99}$ over 500 frames. This satisfies a 33.3\,ms selector-stage deadline but slightly exceeds 15\,ms at $p_{99}$. Because no DPR-specific closure/backend trace is available, we do not report this selector-only measurement as full end-to-end QoS. DPR was exported to an RKNN FP16 artifact, but the RK3588 runtime
probe failed because the current toolchain does not support the
\texttt{Erf} operator used by the model. We therefore report no
RK3588 latency or QoS for DPR and make no runtime extrapolation.

\subsection{Cross-Platform QoS Trade-offs}
\label{subsec:cross_platform_qos}

\begin{figure}[t]
    \centering
    \includegraphics[
    width=0.44\textwidth,
    trim=0 6 0 6,
    clip
    ]{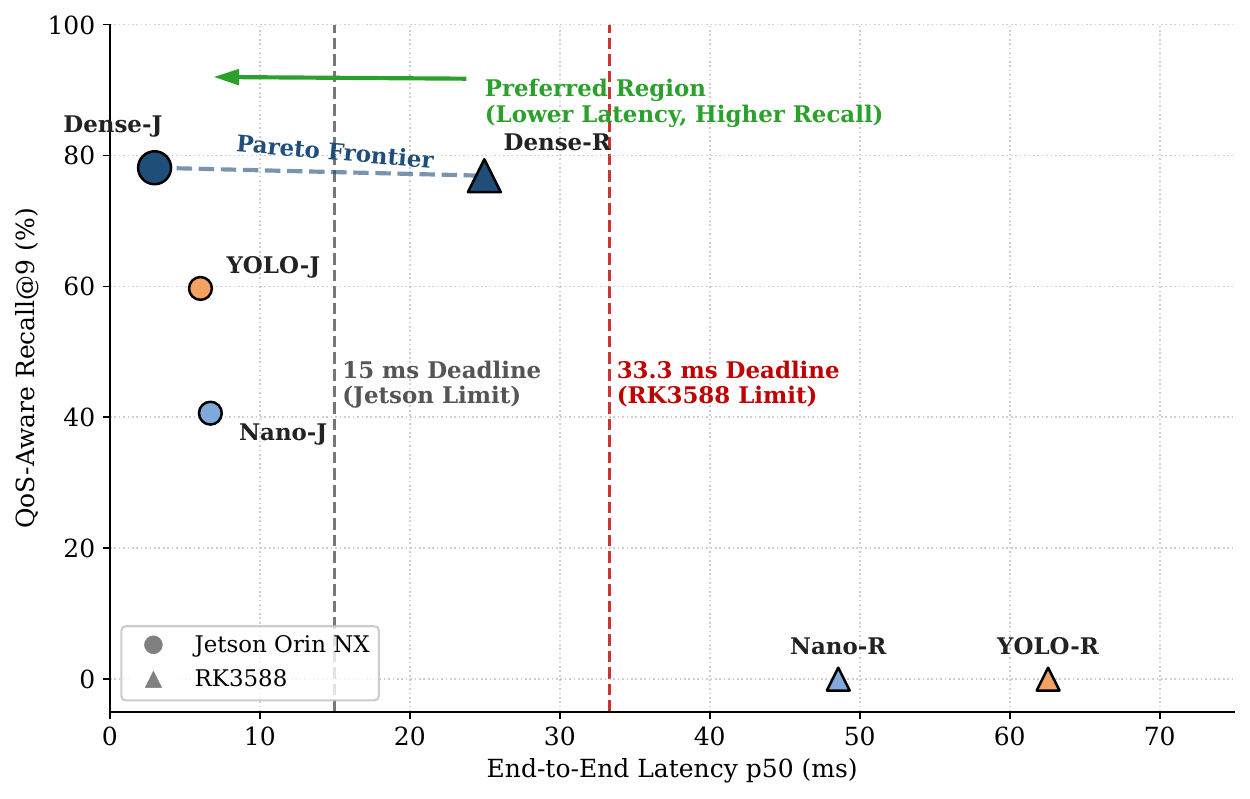}
    \caption{Cross-platform latency--QoS operating points at $K=9$. Each point uses the Copy-Avoidance result reported in Table~\ref{tab:cross_platform_qos_runtime}. DenseScout provides the
best operating point among the evaluated frontends on both Jetson Orin NX and RK3588, under the 15\,ms and 33.3\,ms deadlines, respectively.}
    \Description{Scatter plot of end-to-end latency versus QoS-aware Recall-at-9 on Jetson Orin NX and RK3588, with 15 ms and 33.3 ms deadline boundaries.}
    \label{fig:pareto_cross_platform_main}
\end{figure}

Figure~\ref{fig:pareto_cross_platform_main} summarizes the
cross-platform latency--QoS operating points. DenseScout provides
the lowest latency and the highest QoS-aware recall among the evaluated
frontends on each platform, showing that its advantage is retained when
end-to-end latency becomes part of the evaluation.

To make the deployment behavior more concrete, Table~\ref{tab:cross_platform_qos_runtime} consolidates cross-platform runtime and QoS-aware recall at the practical budget $K=9$. Besides raw and deadline-aware recall, we also report $p50$, $p99$, and tail jitter. On Jetson Orin NX, DenseScout preserves its raw recall almost perfectly under both 33.3\,ms and 15\,ms deadlines, while maintaining the smallest latency and one of the lowest jitter values. On RK3588, DenseScout is the only compared method that remains usable under the 33.3\,ms runtime budget, and it also exhibits dramatically smaller tail jitter than detector-style baselines.

DenseScout is particularly robust under stricter deadlines. Competing detector-style frontends either degrade under tighter latency budgets or fall outside the feasible runtime envelope on constrained hardware. More importantly, the advantage of DenseScout is not limited to lower average latency: it also shows markedly tighter tail behavior, especially on RK3588, which is critical for deadline-bounded edge deployment.

\subsection{Latency Breakdown and Copy-Avoidance Benefit}
\label{subsec:latency_breakdown}

\begin{figure}[!t]
    \centering
    \includegraphics[width=0.45\textwidth]{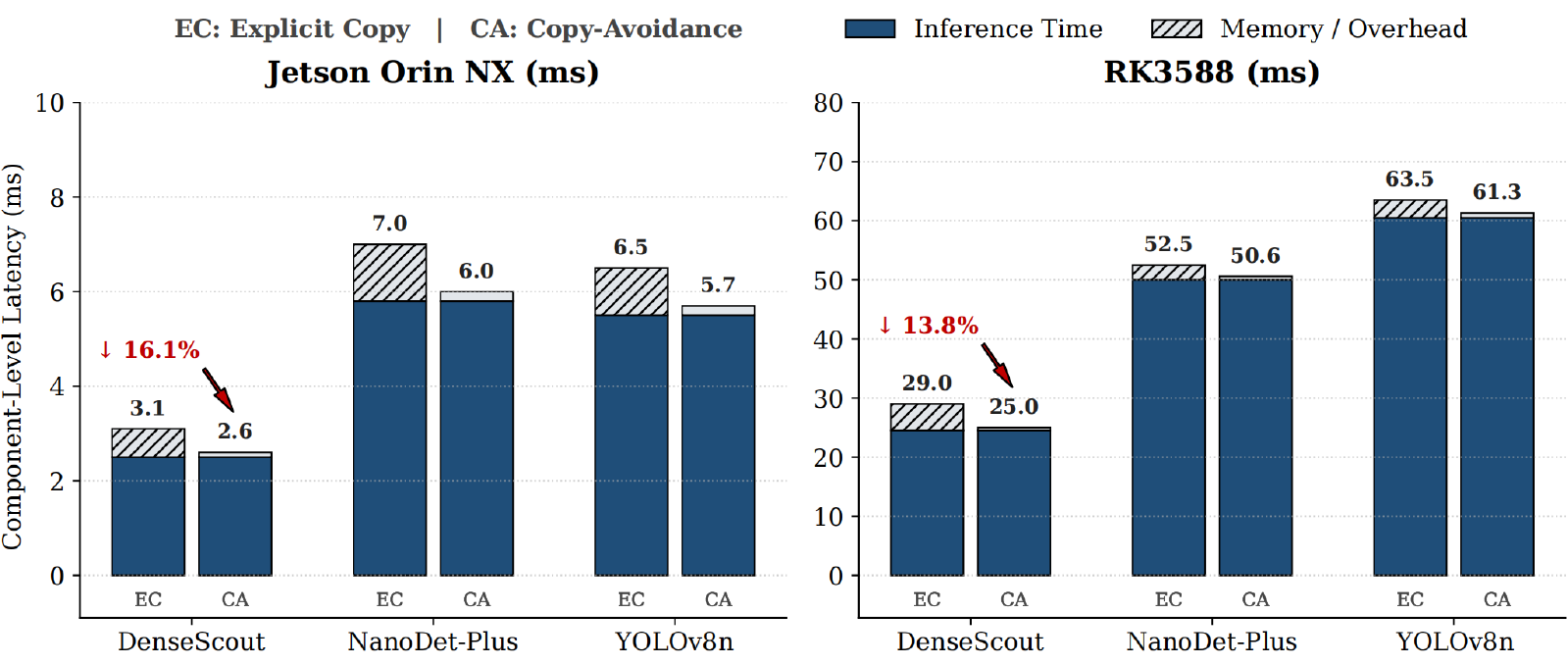}
    \caption{Representative component-level latency decomposition on
Jetson Orin NX and RK3588 under explicit copy (EC) and
Copy-Avoidance (CA). The bars illustrate inference and
memory/transport composition and are not the final per-frame
$p50/p99$ statistics used for QoS computation; those results are
reported in Table~\ref{tab:cross_platform_qos_runtime}.}
\Description{Bar charts showing representative component-level
inference and memory or transport overhead for DenseScout,
NanoDet-Plus, and YOLOv8n on Jetson and RK3588.}
    \label{fig:latency_breakdown_main}
\end{figure}

Figure~\ref{fig:latency_breakdown_main} presents a representative
component-level decomposition of inference and memory/transport
overhead under explicit copy (EC) and Copy-Avoidance (CA).
It is used to explain latency composition rather than to reproduce
the final per-frame latency distribution. DenseScout shows the
lowest total component-level cost on both platforms.

\subsection{Runtime Summary Across Streams and Transport Settings}
\label{subsec:runtime_summary}

To further stress the runtime behavior, Table~\ref{tab:jetson_multistream_qos} reports single-stream and three-stream Jetson Orin NX results at $K=9$. DenseScout maintains low tail latency and nearly unchanged QoS-aware recall across transport settings, whereas NanoDet-Plus exhibits clear degradation once the deadline becomes tight under three-stream concurrency.

\begin{table}[!t]
\centering
\caption{Jetson Orin NX single-stream and three-stream QoS evaluation at $K=9$. Tail behavior is summarized by $p50/p99$ latency.}
\label{tab:jetson_multistream_qos}
\setlength{\tabcolsep}{2.6pt}
\renewcommand{\arraystretch}{0.80}
\footnotesize
\begin{tabular}{llccc}
\toprule
Method & Setting & Raw R@9 & QoS@15ms & p50 / p99 (ms) \\
\midrule
DenseScout   & 1-stream Copy      & 78.12 & 77.91 & 3.08 / 3.16 \\
DenseScout   & 1-stream Copy-Avoidance & 78.12 & 78.12 & 3.01 / 3.09 \\
DenseScout   & 3-stream Copy      & 78.12 & 78.12 & 3.21 / 6.96 \\
DenseScout   & 3-stream Copy-Avoidance & 78.12 & 78.12 & 3.13 / 6.50 \\
\midrule
YOLOv8n      & 1-stream Copy      & 59.68 & 58.97 & 6.20 / 6.32 \\
YOLOv8n      & 1-stream Copy-Avoidance & 59.68 & 59.68 & 6.04 / 6.14 \\
\midrule
NanoDet-Plus & 1-stream Copy      & 40.63 & 40.43 & 6.80 / 6.88 \\
NanoDet-Plus & 1-stream Copy-Avoidance & 40.63 & 40.63 & 6.70 / 6.77 \\
NanoDet-Plus & 3-stream Copy      & 40.63 & 33.54 & 10.45 / 17.46 \\
NanoDet-Plus & 3-stream Copy-Avoidance & 40.63 & 32.83 & 10.72 / 17.43 \\
\bottomrule
\end{tabular}
\end{table}

\begin{figure}[!t]
    \centering
    \includegraphics[width=0.42\textwidth]{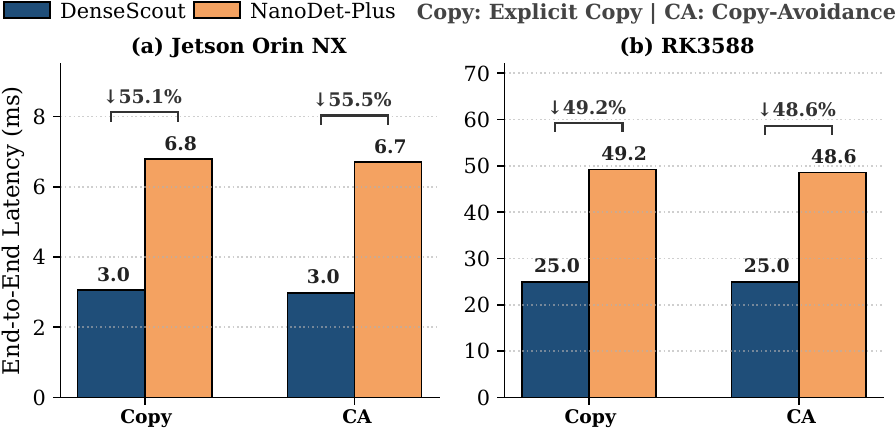}

\caption{End-to-end latency at $K=9$ on Jetson Orin NX and RK3588.
DenseScout is faster than NanoDet-Plus under both transport settings;
deployment QoS is reported in Table~\ref{tab:cross_platform_qos_runtime}.}
    \Description{Two bar charts comparing DenseScout and NanoDet-Plus end-to-end latency under copy and Copy-Avoidance execution on Jetson Orin NX and RK3588.}
    \label{fig:latency_halved_compact}
\end{figure}

For completeness, Figure~\ref{fig:latency_halved_compact} reports representative physical runtime measurements on Jetson Orin NX and RK3588. DenseScout consistently achieves substantially lower end-to-end latency than NanoDet-Plus under both Copy and Copy-Avoidance transport, reducing runtime by 48.6\%--55.5\% across platforms. Additional RK3588 probes across $K \in \{1,4,9,16\}$ show negligible variation, confirming that runtime is effectively insensitive to the tested budget range.

Overall, the deployment results are consistent with the offline findings: DenseScout is not only more effective at covering tiny targets under budget constraints, but also better aligned with real edge-system requirements in terms of latency, deadline validity, and runtime stability.

\section{Conclusion}
\label{sec:conclusion}

We studied budgeted tiny-object selection as a deployment-facing frontend task, distinct from downstream recognition. DenseScout is a compact realization of a ranked patch-center selection interface rather than a new generic detector architecture. Under unified protocols, it provides strong low-budget coverage on VisDrone, DOTA, and InsPLAD, and remains ahead of the evaluated detector-derived and selection-style proxy baselines.

The hardware results further show that offline selector quality is only one component of deployable utility. TensorRT/CUDA execution on Jetson and RKNN/RGA/DMA execution on RK3588 expose different memory-movement and synchronization constraints, yet DenseScout retains a favorable latency--coverage operating point. These results support jointly evaluating selector design, transport-aware realization, and deadline-bounded QoS.


\begin{acks}
This work was supported by the Science and Technology Project of
China Southern Power Grid under Grant No.~GDKJXM20240116
(030000KC24010016).
\end{acks}

\bibliographystyle{ACM-Reference-Format}
\bibliography{references}


\clearpage

\setcounter{section}{0}
\setcounter{subsection}{0}
\setcounter{figure}{0}
\setcounter{table}{0}
\setcounter{equation}{0}

\begin{center}
{\Large\bfseries Supplementary Material for}\\[4pt]
{\large\bfseries
DenseScout: Algorithm-System Co-design for Budgeted Tiny Object Selection on Edge Platforms}
\end{center}

\paragraph{Code availability.}
The source-only alpha implementation of DenseScout is publicly available at
\url{https://github.com/zhouer-bear/DenseScout}.
The current release provides the core model, decoding logic,
public-data utilities, and evaluation interfaces, but does not include
a paper-exact checkpoint, the proprietary Lab\_Dataset, or the complete
vendor-specific board-level deployment wrappers.

\section{Supplementary Overview}

This supplementary document provides additional experimental details, protocol clarifications, extended ablation results, and deployment-oriented profiling data that complement the main paper. The goal is to expose the broader evidence behind the claims in the paper without changing the main narrative. In particular, we include: 
(1) selector-interface and hit-criterion details, 
(2) additional offline budget-sweep results, 
(3) supplementary ablation results,
(4) RK3588 preprocessing and deployment profiling, 
(5) extended Jetson multi-stream QoS data, and 
(6) supplementary seen/unseen generalization results and qualitative evidence.

\begin{figure*}[t]
\centering
\includegraphics[width=0.98\linewidth]{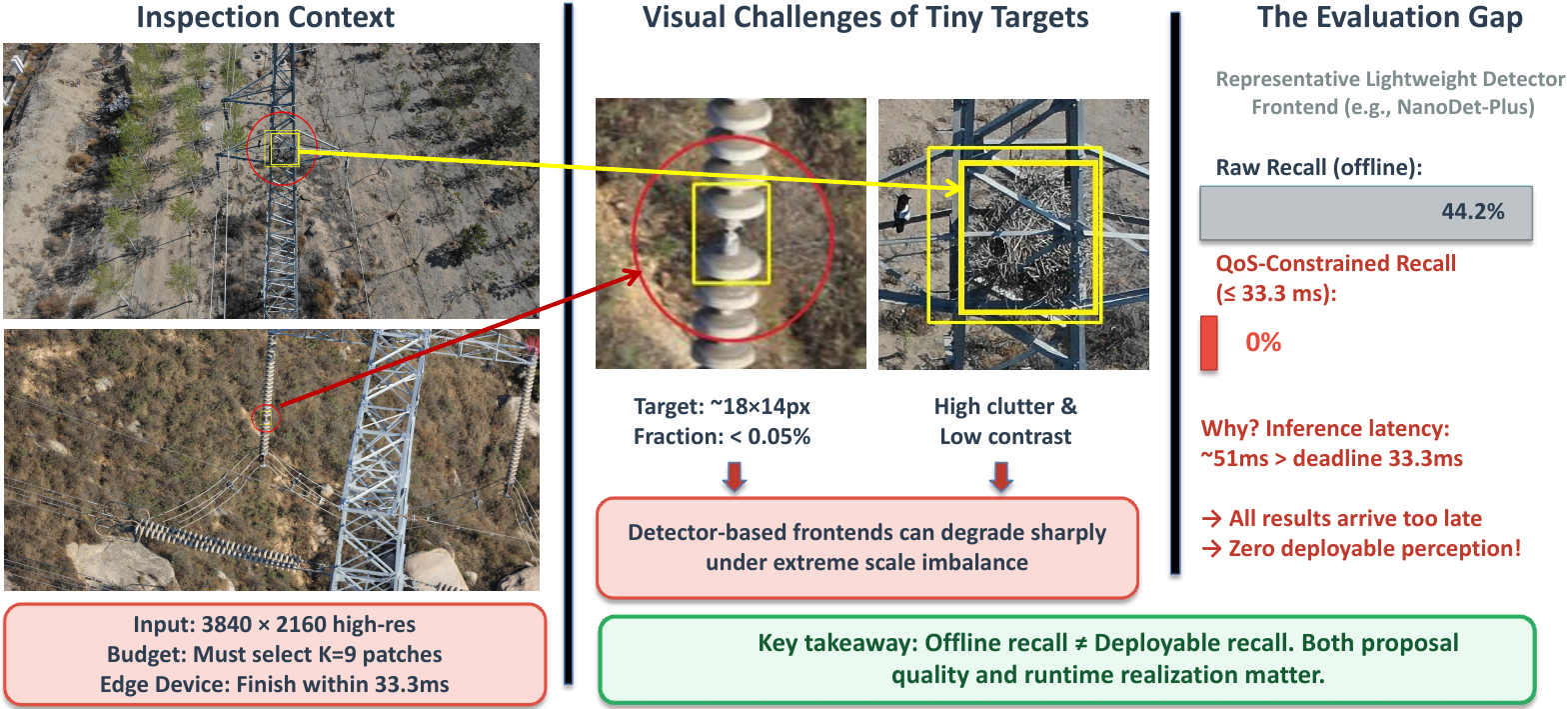}
\caption{Motivating example for deployable tiny-object selection on edge devices. The left panel shows the inspection context under a 4K input and a strict patch budget. The middle panel highlights the extreme scale imbalance, clutter, and low contrast faced by tiny targets. The right panel illustrates the evaluation gap: a lightweight detector-style frontend can still obtain non-trivial offline raw recall, yet deliver zero useful recall under a hard latency deadline when all results arrive too late. This motivates evaluating both proposal quality and runtime-valid deployability.}
\label{fig:supp_hardcore_motivation}
\end{figure*}

\begin{figure*}[t]
\centering
\includegraphics[width=0.95\linewidth]{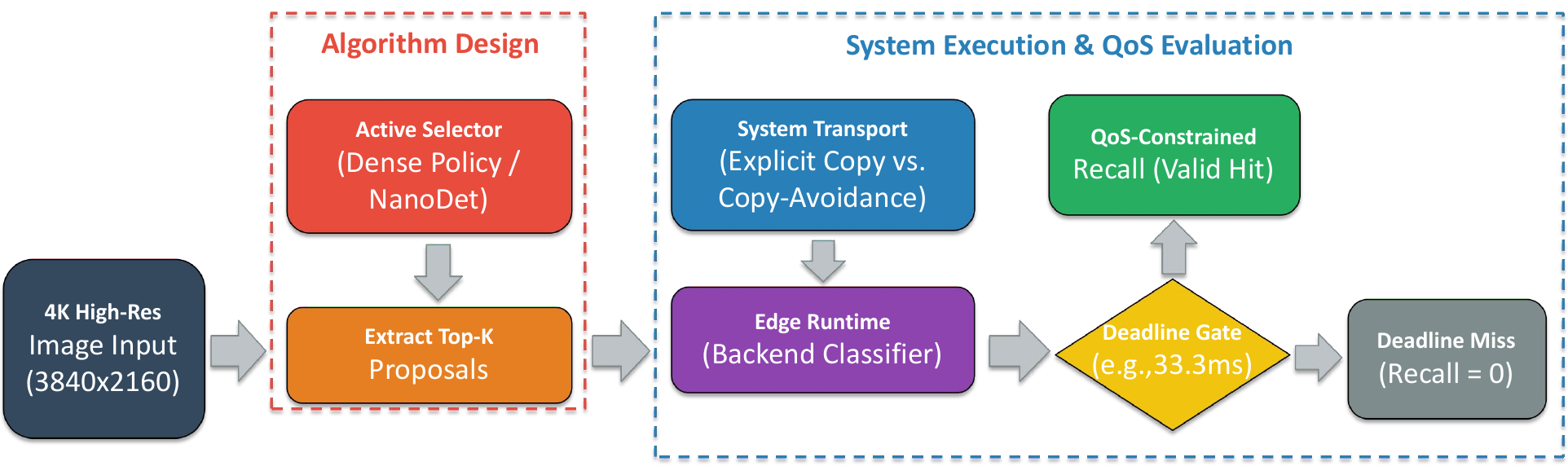}
\caption{Supplementary system schematic used throughout this document. A 4K high-resolution input is first processed by an active selector to extract top-$K$ candidate patches. The system is then evaluated under explicit transport settings and a deadline gate, so that only latency-valid hits contribute to QoS-constrained recall. This figure clarifies the algorithm--system decomposition underlying the supplementary profiling results.}
\label{fig:supp_pipeline_structure}
\end{figure*}

\section{Evaluation Protocol Details}

This section fixes the evaluation protocol used throughout our supplementary experiments. Some legacy supplementary figures and tables use
\emph{Dense Policy} or \texttt{Dense\_Policy} as an earlier
implementation label; both refer to the same DenseScout selector
reported in the main paper.

\subsection{Private Lab Inspection Dataset}

The fixed-$K$ selection-style comparison and the associated
ablation experiments use a proprietary laboratory inspection
dataset. The evaluated subset contains 371 images and 975
point-annotated tiny targets. This dataset is independent of
VisDrone, DOTA, and InsPLAD, and none of its samples contribute
to the public-benchmark results reported in the main paper.

The raw images and location-related metadata cannot be publicly
released because the dataset contains defense-related and
geographically sensitive inspection information. We therefore
disclose the evaluation subset size, annotation type, unified
hit criterion, budget protocol, and aggregate experimental results,
while withholding the original imagery and sensitive metadata.

\subsection{Unified Selector Interface}

To compare dense-response selectors and detector-style frontends under a common patch-budget setting, all methods are converted into a ranked sequence of patch centers:
\begin{itemize}[leftmargin=1.5em]
    \item For DenseScout, local maxima on the dense heatmap are extracted and ranked by response score.
    \item For detector-style frontends, each predicted bounding box is converted to its geometric center, and the resulting centers are ranked by confidence score.
    \item A common Top-$K$ truncation is then applied.
\end{itemize}

This interface ensures that all methods are evaluated as \emph{frontend patch selectors} rather than as full-image detectors.

\subsection{Selection-Style Proxy Baselines}

MaskProposer, FocusGrid, AdaZoom-lite, and GigaDet-DH-lite are
controlled proxy implementations rather than official
reproductions of the corresponding full systems. They instantiate
four broad proposal mechanisms: mask-based proposal scoring,
focus-grid selection, adaptive zoom selection, and
dynamic-head-style proposal scoring, respectively.

All four proxies are converted to the same ranked Top-$K$
patch-center interface used by DenseScout and are evaluated with
the same fixed-$K$ target-level hit criterion. Their purpose is to
test whether these broad proposal mechanisms alone explain the
observed gain under a controlled selection protocol. The reported
results should therefore not be interpreted as complete
reimplementations or full-system comparisons with the original
methods.

\subsection{Target-Level Hit Criterion}

A ground-truth tiny target is counted as covered if at least one selected patch center falls within a fixed spatial neighborhood around the target center. In the offline selector evaluation, the effective coverage cell is a fixed $64\times64$ cell centered at the selected location. Equivalently, a target centered at $(x_g, y_g)$ is considered covered by a selected center $(x_i, y_i)$ if
\begin{equation}
|x_i-x_g| \le 32, \qquad |y_i-y_g| \le 32.
\end{equation}

For some supplementary protocol reports, the same square criterion
is implemented using coordinate-wise center-offset thresholds.

\subsection{Dynamic Effective Budget}

Let $K$ denote the nominal budget and let $K_{\mathrm{eff}}$ denote the number of proposals that survive score filtering and non-maximum suppression:
\begin{equation}
K_{\mathrm{eff}} = \min(K, N_{\mathrm{proposals\ after\ suppression}}).
\end{equation}
In profiling-oriented experiments, the realized system cost may depend on $K_{\mathrm{eff}}$ rather than the nominal $K$.

\subsection{Budget Ratio and Patch-Budget Mapping}

In offline selector evaluation, each selected center corresponds to a fixed square patch on the original image plane. Recall@K is computed by truncating the ranked selector output to the first $K$ retained centers. Recall@Ratio instead evaluates the same ranked sequence under an area budget, where the effective patch ratio is defined as the total selected patch area divided by the full-image area.

Unless otherwise specified, all methods are converted to a common ranked patch-center interface before either Recall@K or Recall@Ratio is computed. Under this protocol, the selector is first sorted by descending confidence (or dense-response score), then truncated according to the requested budget. This ensures that DenseScout and detector-style frontends are compared under the same budget semantics rather than under their native output formats.

\subsection{Patch-Based Detection Closure Protocol}

For the patch-based detection closure benchmark, each frontend first outputs a ranked list of candidate centers under a fixed budget of $K=9$. Each retained center is then expanded to a $640\times640$ backend crop on the original high-resolution image. The same frozen YOLOv8n backend is applied to all frontend-generated crops, and the predicted boxes are mapped back to the global image coordinate system before merging.

To ensure fairness, the backend detector, confidence thresholds, and post-processing settings are shared across all compared frontends. Final predictions from all selected crops are merged by class-aware global non-maximum suppression. Therefore, the closure benchmark isolates the effect of frontend patch selection quality rather than backend detector differences.

For clarity, two spatial units are used in this work. Offline selector recall is computed using a fixed $64\times64$ evaluation cell centered at each selected location, which supports unified Recall@K and Recall@Ratio benchmarking. In closure and deployment experiments, the same selected center is expanded to a larger $640\times640$ backend crop so that downstream patch-level detection preserves sufficient visual context. This distinction is consistent with the main paper and is restated here to avoid ambiguity across supplementary tables and figures.

\section{Additional Offline Results}

\subsection{Supplementary Selector Baselines on a Unified Budget Protocol}

\begin{figure}[t]
\centering
\includegraphics[width=\columnwidth]{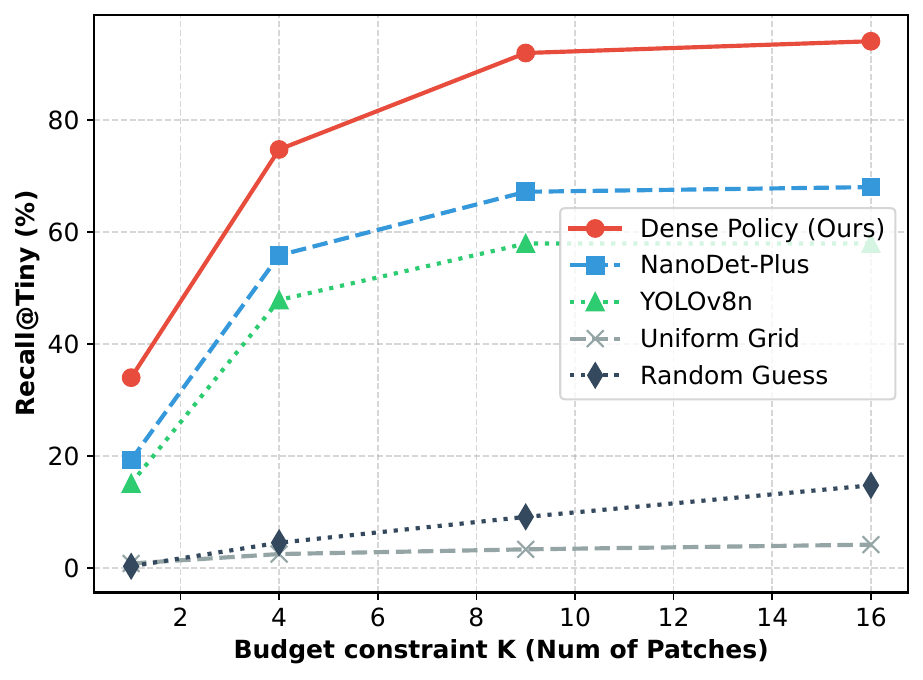}
\caption{Budget--recall comparison under the unified selector protocol. DenseScout maintains a clear advantage over detector-style and heuristic baselines across the tested patch budgets, with the largest practical benefit appearing in the low-budget regime.}
\label{fig:supp_budget_curve}
\end{figure}

Table~\ref{tab:supp_selector_baselines_server} provides an additional selector-oriented comparison with multiple baselines under a common Top-$K$ patch selection protocol on a server-side evaluation split. Fig.~\ref{fig:supp_budget_curve} complements the table by showing the corresponding budget--recall trend.

\begin{table*}[t]
\centering
\caption{Additional selector comparison under a unified Top-$K$ patch-budget protocol. Hit Efficiency is defined as Recall@Tiny$/K$.}
\label{tab:supp_selector_baselines_server}
\setlength{\tabcolsep}{4.0pt}
\renewcommand{\arraystretch}{1.10}
\small
\begin{tabular}{llcccccccc}
\toprule
\multirow{2}{*}{Category} & \multirow{2}{*}{Method}
& \multicolumn{4}{c}{Recall@Tiny (\%)} & \multicolumn{4}{c}{Hit Efficiency (\% / patch)} \\
\cmidrule(lr){3-6} \cmidrule(lr){7-10}
& & K=1 & K=4 & K=9 & K=16 & K=1 & K=4 & K=9 & K=16 \\
\midrule
Active selector & DenseScout / Dense\_Policy & 32.02 & 63.53 & 79.13 & 85.92 & 32.02 & 15.88 & 8.79 & 5.37 \\
Anchor-free detector & NanoDet-Plus & 18.24 & 35.66 & 41.54 & 43.97 & 18.24 & 8.92 & 4.62 & 2.75 \\
Generic detector & YOLOv8n & 16.01 & 31.71 & 37.49 & 39.82 & 16.01 & 7.93 & 4.17 & 2.49 \\
Geometric baseline & Uniform Grid & 1.82 & 4.86 & 8.11 & 11.45 & 1.82 & 1.21 & 0.90 & 0.72 \\
Random baseline & Random Guess & 1.01 & 3.44 & 7.50 & 16.01 & 1.01 & 0.86 & 0.83 & 1.00 \\
\bottomrule
\end{tabular}
\end{table*}

\subsection{Size-Stratified Recall Breakdown}

We additionally report stratified recall under different target-size groups. This supplementary view is useful because the main paper focuses on overall tiny-target selection performance, while the stratified view shows how the selector behaves across different object scales.

\begin{table}[t]
\centering
\caption{Size-stratified Recall@Tiny (\%) under different budgets. Tiny, Small, and Large refer to progressively larger projected target groups in the aligned evaluation coordinate system.}
\label{tab:supp_size_stratified}
\setlength{\tabcolsep}{4.0pt}
\renewcommand{\arraystretch}{1.05}
\small
\begin{tabular}{llcccc}
\toprule
Group & Method & K=1 & K=4 & K=9 & K=16 \\
\midrule
\multirow{3}{*}{Tiny}
& DenseScout & 34.03 & 74.79 & 92.02 & 94.12 \\
& NanoDet-Plus & 19.33 & 55.88 & 67.23 & 68.07 \\
& YOLOv8n & 15.13 & 47.90 & 57.98 & 57.98 \\
\midrule
\multirow{3}{*}{Small}
& DenseScout & 32.46 & 65.10 & 78.80 & 84.43 \\
& NanoDet-Plus & 18.20 & 30.96 & 34.90 & 37.90 \\
& YOLOv8n & 16.89 & 27.95 & 32.46 & 35.27 \\
\midrule
\multirow{3}{*}{Large}
& DenseScout & 28.70 & 47.22 & 65.74 & 80.56 \\
& NanoDet-Plus & 17.13 & 25.00 & 29.63 & 32.41 \\
& YOLOv8n & 14.81 & 23.15 & 27.31 & 31.02 \\
\bottomrule
\end{tabular}
\end{table}

\begin{figure*}[t]
\centering
\includegraphics[width=0.95\linewidth]{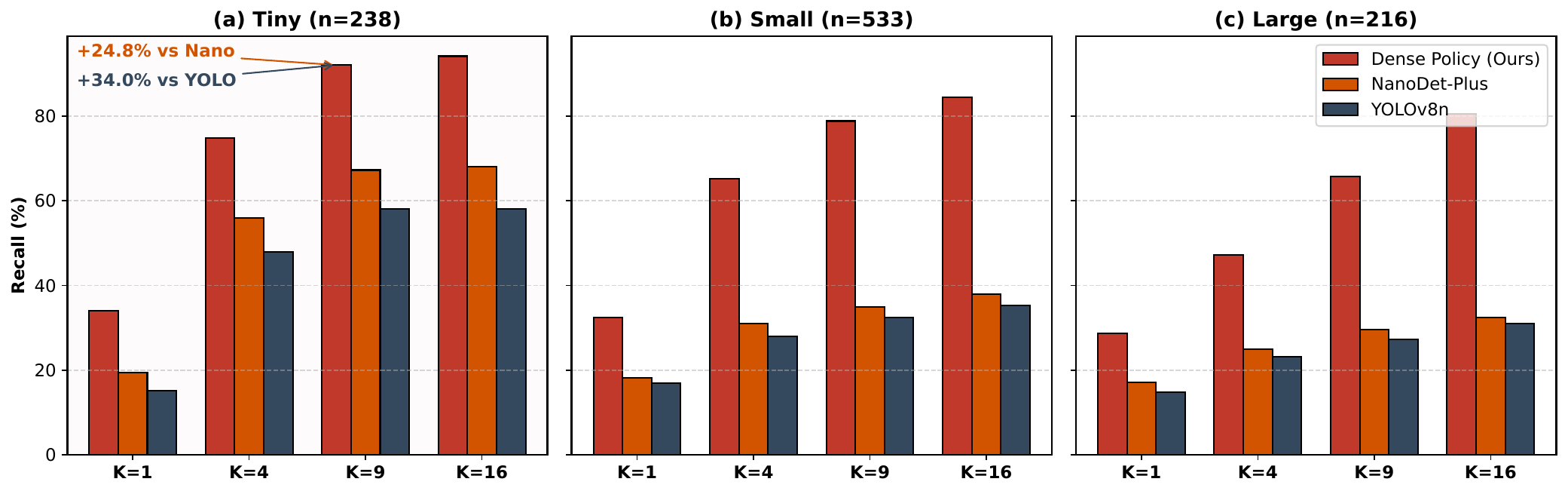}
\caption{Size-stratified recall comparison across different budgets. The three panels report results on tiny, small, and large target groups, respectively. DenseScout shows the strongest relative advantage on the smallest targets, which are also the most challenging and deployment-critical cases in high-resolution inspection.}
\label{fig:supp_size_stratified_fig}
\end{figure*}

\subsection{Supplementary Seen/Unseen Generalization}

\begin{figure*}[t]
\centering
\includegraphics[width=0.95\linewidth]{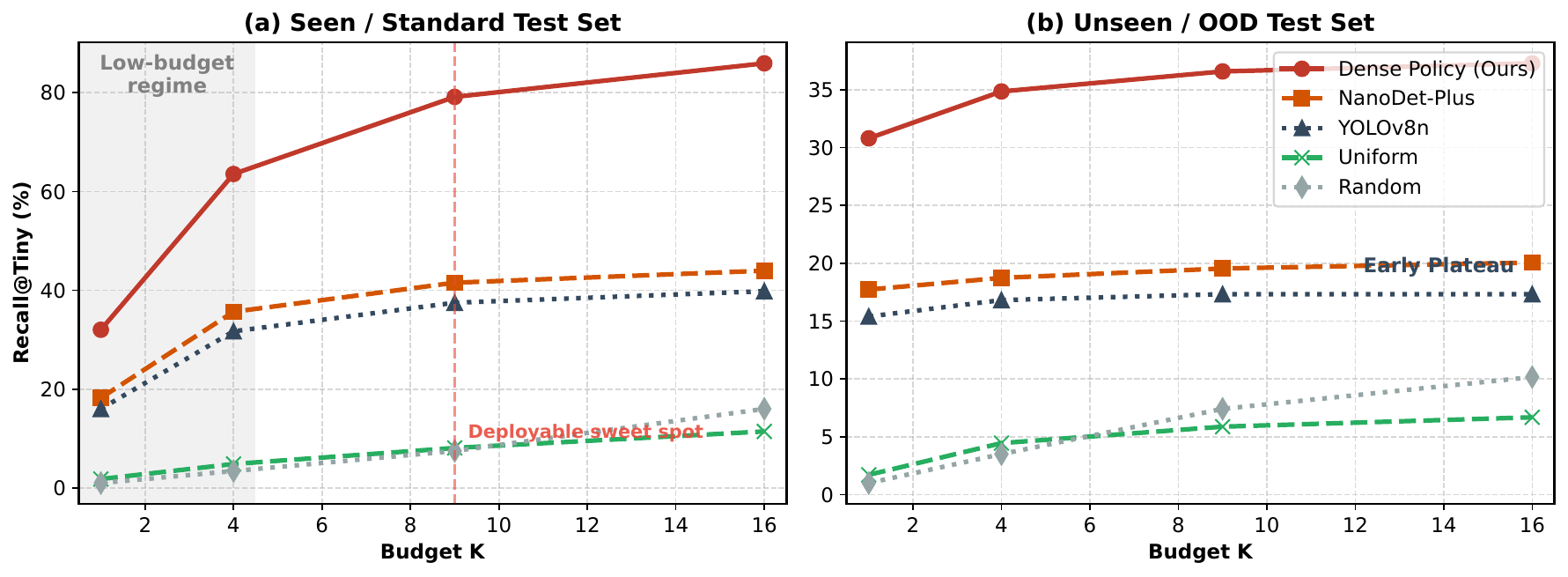}
\caption{Seen versus unseen budget sweep under a unified selector protocol. The seen split highlights the low-budget regime and the practical deployment sweet spot, while the unseen split shows that detector-style baselines plateau earlier under distribution shift. DenseScout remains consistently stronger across both settings.}
\label{fig:supp_seen_unseen}
\end{figure*}

Table~\ref{tab:supp_unseen_generalization} and Fig.~\ref{fig:supp_seen_unseen} provide additional seen/unseen generalization evidence under a unified Top-$K$ patch-budget protocol. These results are supplementary to the main paper and are included to illustrate that the selector ranking behavior remains favorable even when the visual appearance shifts significantly.

\begin{table*}[t]
\centering
\caption{Supplementary unseen-category Recall@Tiny (\%) and hit efficiency under a unified Top-$K$ patch-budget protocol.}
\label{tab:supp_unseen_generalization}
\setlength{\tabcolsep}{4.0pt}
\renewcommand{\arraystretch}{1.10}
\small
\begin{tabular}{llccccc}
\toprule
Method & Metric & K=1 & K=4 & K=9 & K=16 \\
\midrule
\multirow{2}{*}{DenseScout}
& Recall@Tiny (\%) & 30.80 & 34.85 & 36.58 & 37.28 \\
& Hit Efficiency (\% / patch) & 30.80 & 8.71 & 4.06 & 2.33 \\
\midrule
\multirow{2}{*}{NanoDet-Plus}
& Recall@Tiny (\%) & 17.73 & 18.74 & 19.55 & 20.06 \\
& Hit Efficiency (\% / patch) & 17.73 & 4.69 & 2.17 & 1.25 \\
\midrule
\multirow{2}{*}{YOLOv8n}
& Recall@Tiny (\%) & 15.40 & 16.82 & 17.33 & 17.33 \\
& Hit Efficiency (\% / patch) & 15.40 & 4.20 & 1.93 & 1.08 \\
\midrule
\multirow{2}{*}{Uniform Grid}
& Recall@Tiny (\%) & 1.72 & 4.46 & 5.88 & 6.69 \\
& Hit Efficiency (\% / patch) & 1.72 & 1.11 & 0.65 & 0.42 \\
\midrule
\multirow{2}{*}{Random Guess}
& Recall@Tiny (\%) & 1.01 & 3.51 & 7.42 & 10.17 \\
& Hit Efficiency (\% / patch) & 1.01 & 0.88 & 0.82 & 0.64 \\
\bottomrule
\end{tabular}
\end{table*}

\subsection{Additional Qualitative Cases}

To complement the quantitative selector comparisons, we include additional qualitative cases that were omitted from the main paper due to space constraints.

\begin{figure*}[t]
\centering
\includegraphics[width=0.95\linewidth]{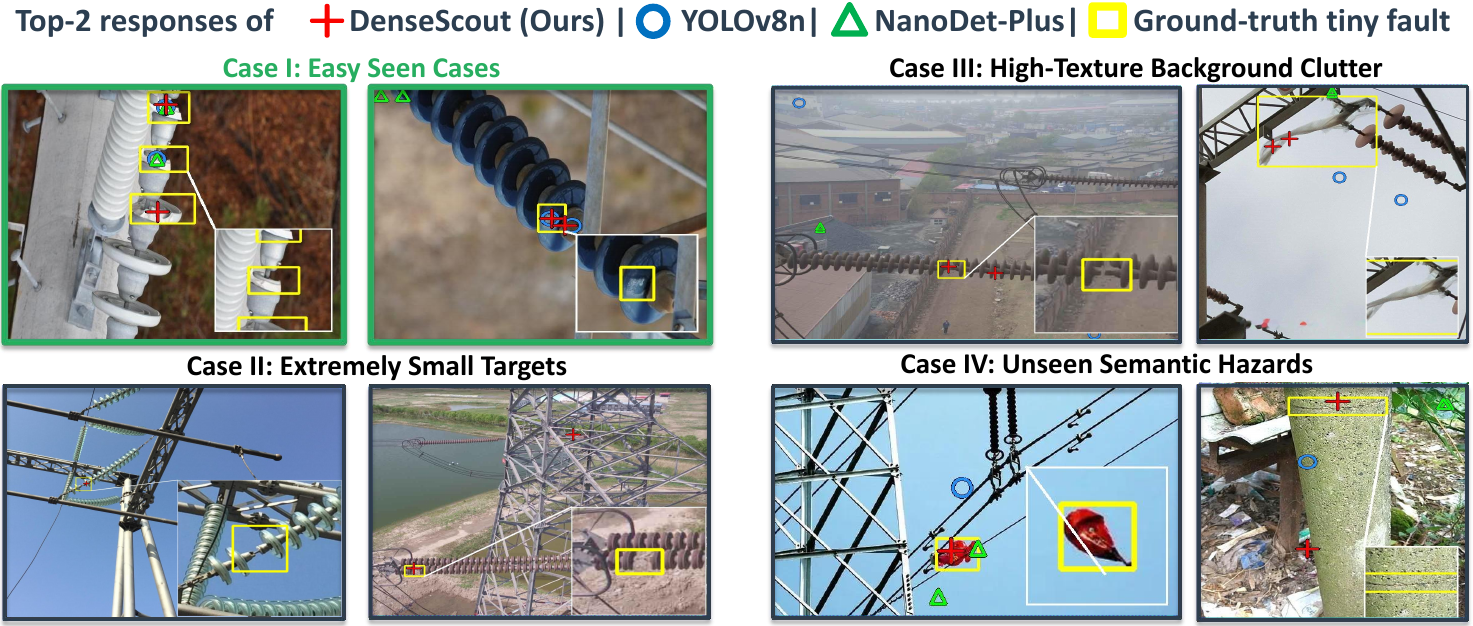}
\caption{Extended qualitative comparison on eight additional cases not shown in the main paper. Yellow boxes indicate ground-truth tiny faults, green triangles denote NanoDet-Plus proposals, red crosses denote DenseScout proposals, and blue circles denote YOLOv8n proposals. The cases cover easy seen examples, extremely small targets, high-texture clutter, and unseen semantic hazards. DenseScout more consistently ranks proposals near the true tiny targets, while detector-style baselines are more easily distracted by large structures or background clutter.}
\label{fig:supp_full_qualitative}
\end{figure*}

\subsection{Orthogonal Ablation of Augmentation and Optimization}

To separate the effect of training augmentation from the effect of optimizer/schedule design, we report the additional orthogonal ablation in Table~\ref{tab:supp_orthogonal_ablation}.

\begin{table}[t]
\centering
\caption{Orthogonal ablation of data augmentation and optimization strategy at $K=9$.}
\label{tab:supp_orthogonal_ablation}
\setlength{\tabcolsep}{1.6pt}
\renewcommand{\arraystretch}{1.08}
\footnotesize
\begin{tabular}{>{\raggedright\arraybackslash}p{2.35cm}cc>{\centering\arraybackslash}p{1.7cm}>{\raggedright\arraybackslash}p{2.45cm}}
\toprule
Variant & Aug. & AdamW+Cosine & Recall@Tiny (\%) & Note \\
\midrule
Tiny-FPN baseline & No & No & 74.15 & Base setting \\
Optimization only & No & Yes & 73.43 & No gain without augmentation \\
Augmentation only & Yes & No & 79.81 & Dominant gain source \\
Full recipe & Yes & Yes & 80.02 & Best overall \\
\bottomrule
\end{tabular}
\end{table}

\section{Additional RK3588 Preprocessing and System Data}

\subsection{4K Resize Preprocessing: CPU vs. RGA vs. DMA-based Copy Avoidance}

Table~\ref{tab:supp_rk3588_resize} reports the 4K preprocessing benchmark that motivated the reduced-copy deployment path.

\begin{table}[t]
\centering
\caption{RK3588 4K resize benchmark from $3840\times2160$ RGB888 to the model input size.}
\label{tab:supp_rk3588_resize}
\setlength{\tabcolsep}{5.0pt}
\renewcommand{\arraystretch}{1.08}
\footnotesize
\begin{tabular}{lccc}
\toprule
Implementation & Avg. latency (ms) & FPS & CPU-visible copy burden \\
\midrule
CPU OpenCV resize & 16.73 & 59.8 & high \\
RGA + malloc & 17.98 & 55.6 & high \\
RGA + DMA buffer & 7.34 & 136.2 & low \\
\bottomrule
\end{tabular}
\end{table}

\subsection{Resolution-Latency Wall on RK3588}

We further include the supplementary resolution sweep that motivates budgeted patch selection instead of direct full-resolution inference.

\begin{table}[t]
\centering
\caption{Resolution-latency wall on RK3588 under computationally matched scaling.}
\label{tab:supp_resolution_wall}
\setlength{\tabcolsep}{5.0pt}
\renewcommand{\arraystretch}{1.08}
\footnotesize
\begin{tabular}{lccc}
\toprule
Resolution & Avg. latency (ms) & FPS & Status \\
\midrule
$640\times640$ & 26.4 & 37.9 & real-time \\
$1280\times720$ & 70.7 & 14.1 & marginal \\
$1920\times1080$ & 167.4 & 6.0 & impractical \\
$3840\times2160$ & 537.4 & 1.9 & unusable \\
\bottomrule
\end{tabular}
\end{table}

\subsection{Naive Zoom Baseline Collapse}

Table~\ref{tab:supp_naive_zoom} reports the serial global-view + patch-view baseline to show why a heavy homogeneous detector pipeline is not a viable frontend design on RK3588.

\begin{table}[t]
\centering
\caption{Naive zoom baseline on RK3588 using a heavy homogeneous detector pipeline. The serial global-view + patch-view design quickly loses real-time viability as the patch budget increases.}
\label{tab:supp_naive_zoom}
\setlength{\tabcolsep}{6.0pt}
\renewcommand{\arraystretch}{1.10}
\footnotesize
\begin{tabular}{cccc}
\toprule
Patch budget $K$ & FPS & Frame time (ms) & Status \\
\midrule
0 & 35.0 & 28.6 & real-time \\
1 & 18.8 & 53.2 & marginal \\
2 & 12.4 & 80.6 & below 15 FPS \\
\bottomrule
\end{tabular}
\end{table}

\subsection{Supplementary RK3588 Single-Core Physical Probe}

Table~\ref{tab:supp_rk3588_singlecore_probe} provides additional physical single-core profiling data. A useful observation is that the frontend runtime is effectively insensitive to the tested $K$ range, which is consistent with the selector design: the main frontend cost is incurred before backend patch processing.

\begin{table}[t]
\centering
\caption{RK3588 single-core physical probe data (median over 100 frames).}
\label{tab:supp_rk3588_singlecore_probe}
\setlength{\tabcolsep}{4.0pt}
\renewcommand{\arraystretch}{1.10}
\small
\begin{tabular}{llcccc}
\toprule
Model & Transport & K=1 & K=4 & K=9 & K=16 \\
\midrule
NanoDet-Plus (FP16) & Copy & 50.23 & 49.24 & 49.34 & 49.23 \\
NanoDet-Plus (FP16) & Copy-Avoidance & 52.19 & 51.39 & 51.39 & 51.38 \\
DenseScout (FP16) & Copy & 24.99 & 25.00 & 24.98 & 24.99 \\
DenseScout (FP16) & Copy-Avoidance & 28.36 & 28.35 & 28.36 & 28.36 \\
\bottomrule
\end{tabular}
\end{table}

\subsection{Supplementary RK3588 Three-Core High-Pressure Probe}

Table~\ref{tab:supp_rk3588_threecore_probe} reports the three-core high-pressure probe for the DenseScout frontend at $K=9$.

\begin{figure}[t]
\centering
\includegraphics[width=0.95\linewidth]{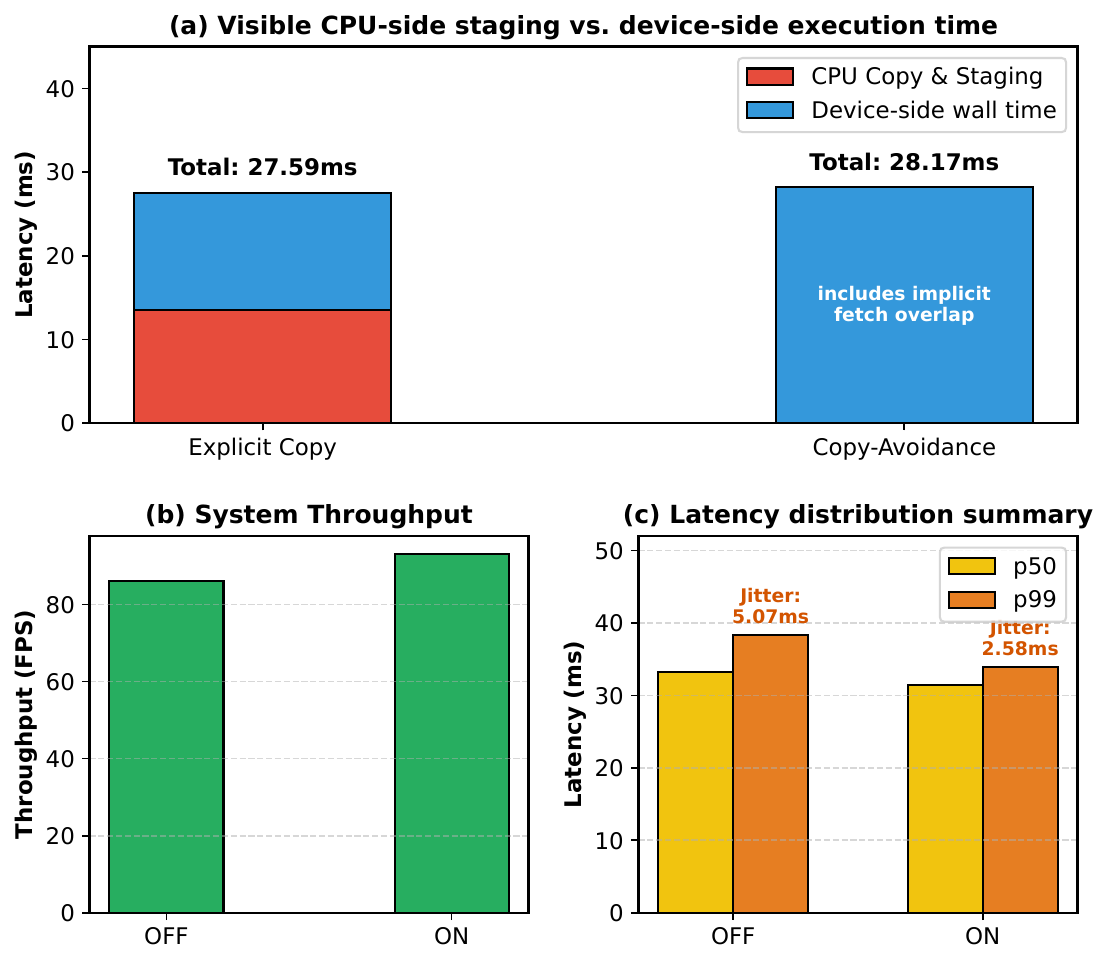}
\caption{RK3588 profiling under explicit copy and copy-avoidance execution modes. (a) Visible CPU-side staging is removed under copy-avoidance, even though device-side wall time remains similar because of implicit overlap. (b) Aggregate system throughput improves when copy overhead is avoided. (c) The latency distribution becomes tighter, with reduced tail jitter under the optimized transport path.}
\label{fig:supp_rk3588_copy_avoidance_fig}
\end{figure}

\begin{table}[t]
\centering
\caption{RK3588 three-core high-pressure profiling for DenseScout at $K=9$.}
\label{tab:supp_rk3588_threecore_probe}
\setlength{\tabcolsep}{5.0pt}
\renewcommand{\arraystretch}{1.08}
\footnotesize
\begin{tabular}{lcccc}
\toprule
Transport & Throughput (FPS) & p50 (ms) & p99 (ms) & Jitter (ms) \\
\midrule
Copy & 98.10 & 28.76 & 31.94 & 3.18 \\
Copy-Avoidance & 88.26 & 31.60 & 34.96 & 3.36 \\
\bottomrule
\end{tabular}
\end{table}

\subsection{Supplementary RK3588 Raw Recall and QoS-Constrained Recall Across Budgets}

Table~\ref{tab:supp_rk3588_qos_budget} exposes the full $K$-sweep on RK3588 used to derive the deployment trends.

\begin{table}[t]
\centering
\caption{Supplementary RK3588 raw and QoS-constrained recall across budgets.}
\label{tab:supp_rk3588_qos_budget}
\setlength{\tabcolsep}{4.0pt}
\renewcommand{\arraystretch}{1.10}
\small
\begin{tabular}{llcccc}
\toprule
Metric & Method & K=1 & K=4 & K=9 & K=16 \\
\midrule
\multirow{2}{*}{Raw Recall@Tiny (\%)}
& NanoDet-Plus & 17.53 & 34.75 & 44.28 & 47.92 \\
& DenseScout & 30.60 & 61.30 & 76.90 & 83.38 \\
\midrule
\multirow{2}{*}{QoS Recall@33.3ms (\%)}
& NanoDet-Plus & 0.00 & 0.00 & 0.00 & 0.00 \\
& DenseScout & 30.60 & 61.30 & 76.90 & 83.38 \\
\midrule
\multirow{2}{*}{QoS Recall@15ms (\%)}
& NanoDet-Plus & 0.00 & 0.00 & 0.00 & 0.00 \\
& DenseScout & 0.00 & 0.00 & 0.00 & 0.00 \\
\bottomrule
\end{tabular}
\end{table}

\begin{figure*}[t]
\centering
\includegraphics[width=0.95\linewidth]{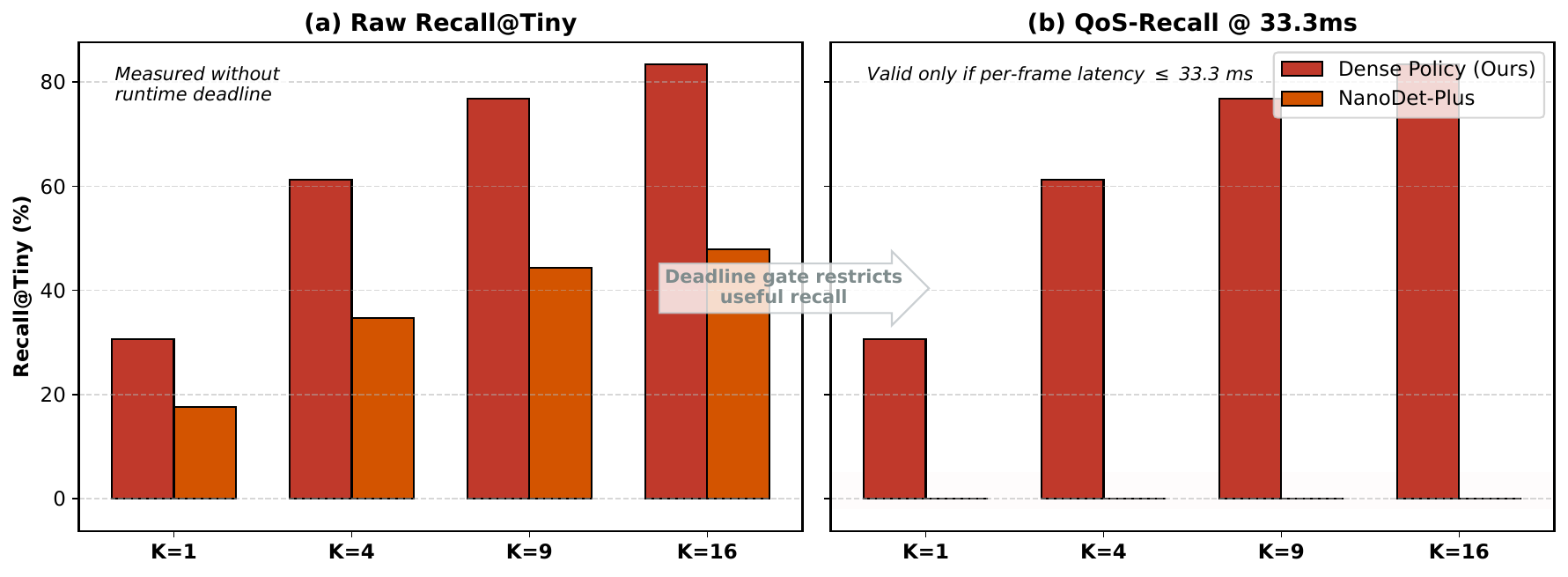}
\caption{Raw recall versus QoS-constrained recall under a 33.3\,ms deadline on RK3588. The left panel reports algorithmic recall without deadline filtering, while the right panel counts only latency-valid outputs. This figure makes explicit that offline recall and deployable recall can diverge sharply under hard real-time constraints.}
\label{fig:supp_raw_vs_qos_fig}
\end{figure*}

\section{Additional Jetson Orin NX Results}

\subsection{Jetson Single- and Multi-Stream QoS Sweep on the Seen Split}

Table~\ref{tab:supp_jetson_seen_compact} provides the full Jetson QoS matrix on the seen split. This extends the smaller table in the main paper by exposing multiple $K$ points and both transport settings.

\begin{table}[t]
\centering
\caption{Compact Jetson Orin NX QoS matrix on the seen split. Latency distribution details are reported separately in Table~\ref{tab:supp_jetson_transport_stress} and Fig.~\ref{fig:supp_jetson_strict_qos_fig}.}

\label{tab:supp_jetson_seen_compact}
\setlength{\tabcolsep}{3.2pt}
\renewcommand{\arraystretch}{1.10}
\scriptsize
\begin{tabular}{llccccc}
\toprule
Method & Transport & Streams & K & Raw Recall (\%) & QoS@33ms (\%) & QoS@15ms (\%) \\
\midrule
DenseScout & Copy & 1 & 1  & 31.61 & 31.61 & 31.61 \\
DenseScout & Copy & 1 & 4  & 63.73 & 63.73 & 63.53 \\
DenseScout & Copy & 1 & 9  & 78.12 & 78.12 & 77.91 \\
DenseScout & Copy & 1 & 16 & 85.61 & 85.61 & 85.41 \\
DenseScout & Copy-Avoidance & 1 & 1  & 31.61 & 31.61 & 31.61 \\
DenseScout & Copy-Avoidance & 1 & 4  & 63.73 & 63.73 & 63.73 \\
DenseScout & Copy-Avoidance & 1 & 9  & 78.12 & 78.12 & 78.12 \\
DenseScout & Copy-Avoidance & 1 & 16 & 85.61 & 85.61 & 85.61 \\
\midrule
NanoDet-Plus & Copy & 1 & 1  & 17.93 & 17.93 & 17.83 \\
NanoDet-Plus & Copy & 1 & 4  & 34.55 & 34.55 & 34.35 \\
NanoDet-Plus & Copy & 1 & 9  & 40.63 & 40.63 & 40.43 \\
NanoDet-Plus & Copy & 1 & 16 & 42.55 & 42.55 & 42.35 \\
NanoDet-Plus & Copy-Avoidance & 1 & 1  & 17.93 & 17.93 & 17.93 \\
NanoDet-Plus & Copy-Avoidance & 1 & 4  & 34.55 & 34.55 & 34.55 \\
NanoDet-Plus & Copy-Avoidance & 1 & 9  & 40.63 & 40.63 & 40.63 \\
NanoDet-Plus & Copy-Avoidance & 1 & 16 & 42.55 & 42.55 & 42.55 \\
\midrule
DenseScout & Copy & 3 & 1  & 31.61 & 31.61 & 31.61 \\
DenseScout & Copy & 3 & 4  & 63.73 & 63.73 & 63.73 \\
DenseScout & Copy & 3 & 9  & 78.12 & 78.12 & 78.12 \\
DenseScout & Copy & 3 & 16 & 85.61 & 85.61 & 85.61 \\
DenseScout & Copy-Avoidance & 3 & 1  & 31.61 & 31.61 & 31.61 \\
DenseScout & Copy-Avoidance & 3 & 4  & 63.73 & 63.73 & 63.73 \\
DenseScout & Copy-Avoidance & 3 & 9  & 78.12 & 78.12 & 78.12 \\
DenseScout & Copy-Avoidance & 3 & 16 & 85.61 & 85.61 & 85.61 \\
\midrule
NanoDet-Plus & Copy & 3 & 1  & 17.93 & 17.93 & 15.20 \\
NanoDet-Plus & Copy & 3 & 4  & 34.55 & 34.55 & 29.08 \\
NanoDet-Plus & Copy & 3 & 9  & 40.63 & 40.63 & 33.54 \\
NanoDet-Plus & Copy & 3 & 16 & 42.55 & 42.55 & 35.36 \\
NanoDet-Plus & Copy-Avoidance & 3 & 1  & 17.93 & 17.93 & 14.69 \\
NanoDet-Plus & Copy-Avoidance & 3 & 4  & 34.55 & 34.55 & 28.17 \\
NanoDet-Plus & Copy-Avoidance & 3 & 9  & 40.63 & 40.63 & 32.83 \\
NanoDet-Plus & Copy-Avoidance & 3 & 16 & 42.55 & 42.55 & 34.75 \\
\bottomrule
\end{tabular}
\end{table}

\subsection{Jetson Single- and Multi-Stream QoS Sweep on the Unseen Split}

Table~\ref{tab:supp_jetson_unseen_compact} reports the corresponding unseen split evaluation.

\begin{table}[t]
\centering
\caption{Compact Jetson Orin NX QoS matrix on the unseen split.}
\label{tab:supp_jetson_unseen_compact}
\setlength{\tabcolsep}{3.2pt}
\renewcommand{\arraystretch}{1.10}
\scriptsize
\begin{tabular}{llccccc}
\toprule
Method & Transport & Streams & K & Raw Recall (\%) & QoS@33ms (\%) & QoS@15ms (\%) \\
\midrule
DenseScout & Copy & 1 & 1  & 7.70 & 7.70 & 7.70 \\
DenseScout & Copy & 1 & 4  & 21.24 & 21.24 & 21.24 \\
DenseScout & Copy & 1 & 9  & 34.55 & 34.55 & 34.55 \\
DenseScout & Copy & 1 & 16 & 45.25 & 45.25 & 45.25 \\
DenseScout & Copy-Avoidance & 1 & 1  & 7.70 & 7.70 & 7.70 \\
DenseScout & Copy-Avoidance & 1 & 4  & 21.24 & 21.24 & 21.24 \\
DenseScout & Copy-Avoidance & 1 & 9  & 34.55 & 34.55 & 34.55 \\
DenseScout & Copy-Avoidance & 1 & 16 & 45.25 & 45.25 & 45.25 \\
\midrule
NanoDet-Plus & Copy & 1 & 1  & 1.81 & 1.81 & 1.81 \\
NanoDet-Plus & Copy & 1 & 4  & 3.81 & 3.81 & 3.81 \\
NanoDet-Plus & Copy & 1 & 9  & 4.85 & 4.85 & 4.85 \\
NanoDet-Plus & Copy & 1 & 16 & 5.29 & 5.29 & 5.29 \\
NanoDet-Plus & Copy-Avoidance & 1 & 1  & 1.81 & 1.81 & 1.81 \\
NanoDet-Plus & Copy-Avoidance & 1 & 4  & 3.81 & 3.81 & 3.81 \\
NanoDet-Plus & Copy-Avoidance & 1 & 9  & 4.85 & 4.85 & 4.85 \\
NanoDet-Plus & Copy-Avoidance & 1 & 16 & 5.29 & 5.29 & 5.29 \\
\midrule
DenseScout & Copy & 3 & 1  & 7.70 & 7.70 & 7.70 \\
DenseScout & Copy & 3 & 4  & 21.24 & 21.24 & 21.24 \\
DenseScout & Copy & 3 & 9  & 34.55 & 34.55 & 34.55 \\
DenseScout & Copy & 3 & 16 & 45.25 & 45.25 & 45.25 \\
DenseScout & Copy-Avoidance & 3 & 1  & 7.70 & 7.70 & 7.70 \\
DenseScout & Copy-Avoidance & 3 & 4  & 21.24 & 21.24 & 21.24 \\
DenseScout & Copy-Avoidance & 3 & 9  & 34.55 & 34.55 & 34.55 \\
DenseScout & Copy-Avoidance & 3 & 16 & 45.25 & 45.25 & 45.25 \\
\midrule
NanoDet-Plus & Copy & 3 & 1  & 1.81 & 1.81 & 1.78 \\
NanoDet-Plus & Copy & 3 & 4  & 3.81 & 3.81 & 3.74 \\
NanoDet-Plus & Copy & 3 & 9  & 4.85 & 4.85 & 4.77 \\
NanoDet-Plus & Copy & 3 & 16 & 5.29 & 5.29 & 5.22 \\
NanoDet-Plus & Copy-Avoidance & 3 & 1  & 1.81 & 1.81 & 1.81 \\
NanoDet-Plus & Copy-Avoidance & 3 & 4  & 3.81 & 3.81 & 3.81 \\
NanoDet-Plus & Copy-Avoidance & 3 & 9  & 4.85 & 4.85 & 4.81 \\
NanoDet-Plus & Copy-Avoidance & 3 & 16 & 5.29 & 5.29 & 5.25 \\
\bottomrule
\end{tabular}
\end{table}

\subsection{Supplementary Jetson Throughput and Tail-Latency Stress Test}

Table~\ref{tab:supp_jetson_transport_stress} includes the additional transport stress test results under more explicit throughput and deadline-miss reporting.

\begin{table}[t]
\centering
\caption{Jetson transport stress test under concurrent execution.}
\label{tab:supp_jetson_transport_stress}
\setlength{\tabcolsep}{3.6pt}
\renewcommand{\arraystretch}{1.08}
\footnotesize
\resizebox{\columnwidth}{!}{%
\begin{tabular}{llccccc}
\toprule
Transport & Streams & Throughput (FPS) & Deadline miss @15ms (\%) & p50 (ms) & p99 (ms) & Jitter (ms) \\
\midrule
Explicit Copy & 1 & 90.0 & 7.67 & 9.30 & 17.68 & 8.38 \\
Copy-Avoidance & 1 & 242.0 & 0.00 & 3.82 & 4.84 & 1.02 \\
Explicit Copy & 3 & 155.2 & 97.56 & 18.92 & 21.49 & 2.57 \\
Copy-Avoidance & 3 & 269.1 & 0.56 & 11.30 & 14.43 & 3.13 \\
\bottomrule
\end{tabular}%
}
\end{table}

\subsection{Supplementary Jetson Time Breakdown}

Table~\ref{tab:supp_jetson_time_breakdown} gives the per-stream average breakdown used to interpret the transport overhead.

\begin{table}[t]
\centering
\caption{Jetson per-stream average time breakdown.}
\label{tab:supp_jetson_time_breakdown}
\setlength{\tabcolsep}{3.0pt}
\renewcommand{\arraystretch}{0.95}
\footnotesize
\begin{tabular}{llccc}
\toprule
Transport & Streams & Host Copy (ms) & GPU Execute (ms) & CPU Sync/Staging (ms) \\
\midrule
Explicit Copy & 1 & 2.57 & 7.34 & 0.44 \\
Copy-Avoidance & 1 & 0.01 & 3.55 & 0.38 \\
Explicit Copy & 3 & 2.82 & 15.32 & 0.50 \\
Copy-Avoidance & 3 & 0.06 & 10.29 & 0.47 \\
\bottomrule
\end{tabular}
\end{table}

\begin{figure}[t]
\centering
\includegraphics[width=0.95\linewidth]{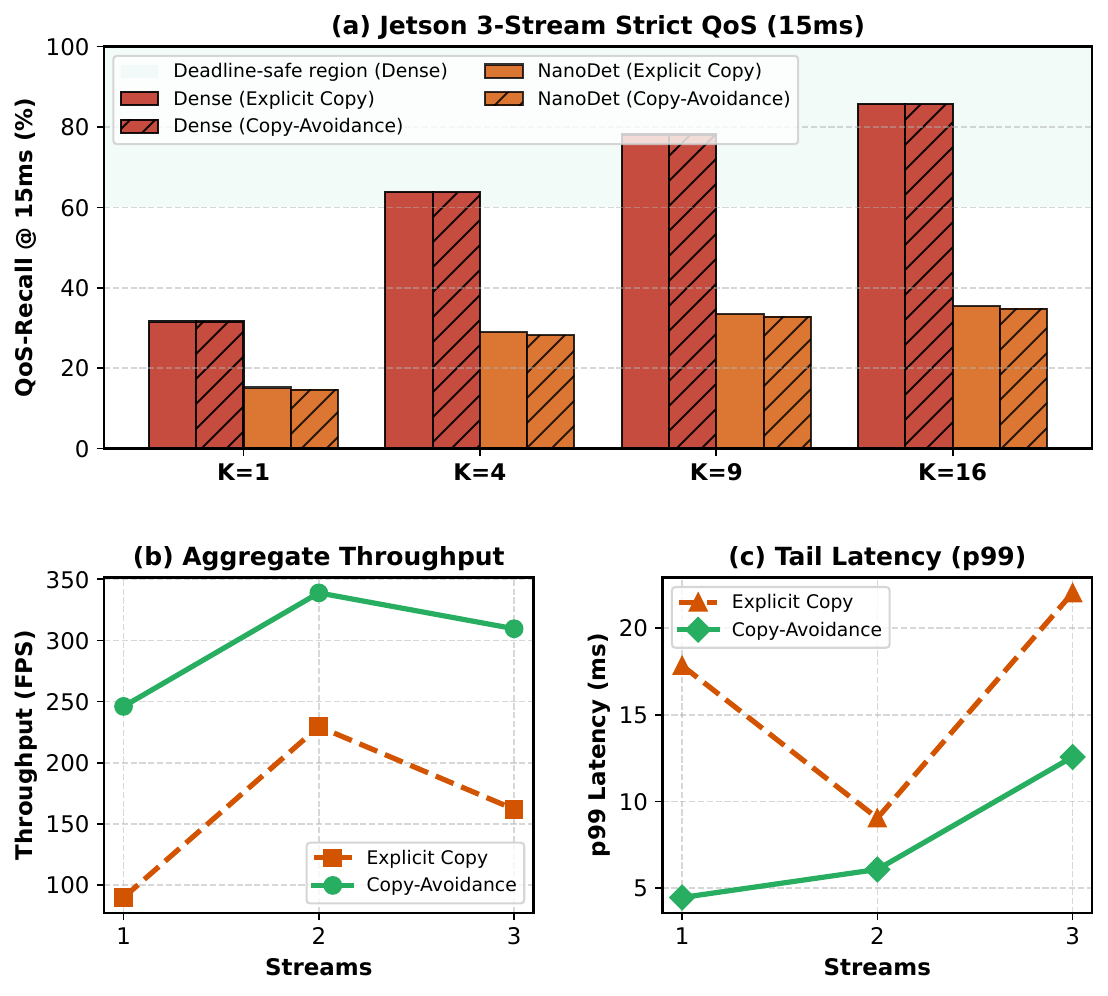}
\caption{Jetson Orin NX profiling under a strict 15\,ms QoS target. The top panel reports QoS-valid recall under 3-stream concurrent execution, while the bottom panels show aggregate throughput and tail latency. DenseScout preserves a large QoS-valid advantage under concurrency, and copy-avoidance further improves system-level robustness.}
\label{fig:supp_jetson_strict_qos_fig}
\end{figure}

\section{Supplementary Notes on Implementation}

\subsection{Training and Reproducibility Notes}

The retained DenseScout training runs were conducted on a shared
server equipped with 8$\times$ NVIDIA A100 GPUs. The training runs
typically lasted approximately 70 epochs, with minor variation
across experiments, and used AdamW with cosine annealing,
random horizontal flipping, and HSV jitter.

These experiments were completed before a standardized per-run
experiment ledger was introduced. Consequently, exact epoch counts,
random seeds, and some optimizer hyperparameters were not
consistently preserved for every historical run. We therefore report
only settings that can be verified from the retained source,
checkpoints, and experimental outputs, rather than reconstructing
missing values retrospectively. Within each controlled comparison,
all methods use the same evaluation split, annotation set, patch
budget, and target-level hit criterion.

\subsection{Platform and Toolchain Notes}

The supplementary experiments were collected across three representative environments: a server platform for training and offline evaluation, an RK3588 edge platform for deployment-oriented profiling, and a Jetson Orin NX platform for multi-stream QoS analysis.

For training and offline evaluation, we used a server equipped with 8$\times$ NVIDIA A100 GPUs. The edge deployment experiments on RK3588 were conducted on a commercial-grade RK3588 core-board platform with 16\,GB RAM and 128\,GB storage. Representative RK3588 deployment toolchains included RKNN-based inference and reduced-copy preprocessing paths using RGA/DMA-backed buffers when applicable.

The Jetson experiments were conducted on an NVIDIA Jetson Orin NX 16\,GB module integrated in a Y-C11 development kit with a 256\,GB SSD. The device exposes dual Gigabit Ethernet, USB 3.0, HDMI, M.2 expansion, and multi-lane MIPI camera interfaces. Representative Jetson deployment used TensorRT-based execution contexts, with independent execution contexts, CUDA streams, and dedicated buffer space assigned to each stream during multi-stream evaluation.

We report hardware-aware latency, throughput, and QoS-constrained recall exactly as measured on the corresponding target devices. Minor software-version differences that do not affect the measurement protocol are not expected to change the qualitative conclusions of the supplementary analysis.

\begin{table}[t]
\centering
\caption{Representative hardware and deployment environments used in the supplementary experiments.}
\label{tab:supp_platforms}
\setlength{\tabcolsep}{3.5pt}
\renewcommand{\arraystretch}{1.08}
\scriptsize
\begin{tabular}{>{\raggedright\arraybackslash}p{1.45cm} >{\raggedright\arraybackslash}p{1.55cm} >{\raggedright\arraybackslash}p{2.55cm} >{\raggedright\arraybackslash}p{2.35cm}}
\toprule
Platform & Role & Hardware & Toolchain \\
\midrule
Server & Training / offline eval. & 8$\times$ NVIDIA A100 GPUs & PyTorch-based training and offline evaluation \\
RK3588 & Edge deployment profiling & Commercial RK3588 core board, 16\,GB RAM, 128\,GB storage, carrier board & RKNN-based inference; RGA / DMA-backed reduced-copy preprocessing \\
Jetson Orin NX & Multi-stream QoS eval. & Jetson Orin NX 16\,GB on Y-C11 kit, 256\,GB SSD & TensorRT-based inference with per-stream execution contexts and CUDA streams \\
\bottomrule
\end{tabular}
\end{table}

\subsection{Scope of Board-Level Deployment Comparison}

DPR is included in the offline selector benchmarking and in the
patch-based closure study because it is a relevant selective
high-resolution baseline. It is not included in the full
end-to-end board-level QoS comparison on RK3588 and Jetson Orin
NX because no DPR-specific closure/backend trace is available
under the unified lightweight runtime path.

As reported in the main paper, we provide a selector-only
TensorRT probe on Jetson Orin NX. We also attempted RKNN FP16
deployment on RK3588, but the runtime probe failed because the
current toolchain does not support the \texttt{Erf} operator used
by the model. We therefore avoid extrapolating full DPR latency or
QoS from incomplete selector-stage measurements.

\section{Closing Remark}

The additional evidence in this supplementary document supports the same central picture as the main paper: DenseScout is not only stronger as a low-budget tiny-object selector in offline evaluation, but also more compatible with deployable edge execution once transport, concurrency, and deadline-aware utility are explicitly accounted for.

\clearpage


\end{document}